\def\year{2022}\relax
\documentclass[letterpaper]{article} % DO NOT CHANGE THIS
\usepackage{aaai22}  % DO NOT CHANGE THIS
\usepackage{times}  % DO NOT CHANGE THIS
\usepackage{helvet}  % DO NOT CHANGE THIS
\usepackage{courier}  % DO NOT CHANGE THIS
\usepackage[hyphens]{url}  % DO NOT CHANGE THIS
\usepackage{graphicx} % DO NOT CHANGE THIS
\usepackage{natbib}  % DO NOT CHANGE THIS AND DO NOT ADD ANY OPTIONS TO IT
\usepackage{caption} % DO NOT CHANGE THIS AND DO NOT ADD ANY OPTIONS TO IT
\DeclareCaptionStyle{ruled}{labelfont=normalfont,labelsep=colon,strut=off} % DO NOT CHANGE THIS
\usepackage{algorithm}
\usepackage{algorithmic}

\usepackage{amssymb,amsfonts}
\usepackage{booktabs}
\usepackage{xcolor}
\usepackage{multirow}
\usepackage{caption}
\usepackage{subcaption}
\usepackage{longtable}
\usepackage{amsmath}
\usepackage{float}
\usepackage{makecell}
\usepackage{listings}
\usepackage{hyperref}

\definecolor{codegreen}{rgb}{0,0.6,0}
\definecolor{codegray}{rgb}{0.5,0.5,0.5}
\definecolor{codepurple}{rgb}{0.58,0,0.82}
\definecolor{backcolour}{rgb}{0.95,0.95,0.92}
\definecolor{elenscolor}{HTML}{029E73}
\definecolor{folcolor}{HTML}{CFBF00}

\lstdefinestyle{mystyle}{
    backgroundcolor=\color{backcolour},   
    commentstyle=\color{codegreen},
    keywordstyle=\color{magenta},
    numberstyle=\tiny\color{codegray},
    stringstyle=\color{codepurple},
    basicstyle=\ttfamily\footnotesize,
    breakatwhitespace=false,         
    breaklines=true,                 
    captionpos=b,                    
    keepspaces=true,                 
    numbers=left,                    
    numbersep=5pt,                  
    showspaces=false,                
    showstringspaces=false,
    showtabs=false,                  
    tabsize=2
}
\newcommand{\pb}[1]{\textcolor{black}{#1}}

\usepackage{newfloat}
\usepackage{listings}
\floatstyle{ruled}
\newfloat{listing}{tb}{lst}{}
\floatname{listing}{Listing}
\title{Entropy-based Logic Explanations of Neural Networks}
\author{Paper ID: 2935}
\title{Entropy-based Logic Explanations of Neural Networks}
\author {
    Pietro Barbiero\textsuperscript{\rm 1}\href{https://orcid.org/0000-0003-3155-2564}{\includegraphics[scale=0.06]{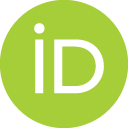}},
    Gabriele Ciravegna \textsuperscript{\rm 2,3,4}\href{https://orcid.org/0000-0002-6799-1043}{\includegraphics[scale=0.06]{figs/orcid.png}},
    Francesco Giannini \textsuperscript{\rm 3}\href{https://orcid.org/ 0000-0001-8492-8110}{\includegraphics[scale=0.06]{figs/orcid.png}},\\
    Pietro Li\'o \textsuperscript{\rm 1}\href{https://orcid.org/ 0000-0002-0540-5053}{\includegraphics[scale=0.06]{figs/orcid.png}},
    Marco Gori \textsuperscript{\rm 3,4}\href{https://orcid.org/0000-0001-6337-5430}{\includegraphics[scale=0.06]{figs/orcid.png}},
    Stefano Melacci \textsuperscript{\rm 3}\href{https://orcid.org/0000-0002-0415-0888}{\includegraphics[scale=0.06]{figs/orcid.png}},
}
\affiliations {
    \textsuperscript{\rm 1} University of Cambridge (UK), 
    \textsuperscript{\rm 2} Università di Firenze (Italy),
    \textsuperscript{\rm 3} Università di Siena (Italy),
    \textsuperscript{\rm 4} Universit\'e Côte d’Azur (France)\\
    \{pb737, pl213\}@cam.ac.uk, gabriele.ciravegna@unifi.it, \{francesco.giannini, marco.gori, mela\}@unisi.it
}
\begin{document}

\maketitle

\begin{abstract}
Explainable artificial intelligence has rapidly emerged since lawmakers have started requiring interpretable models for safety-critical domains. Concept-based neural networks have arisen as explainable-by-design methods as they leverage human-understandable symbols (i.e. concepts) to predict class memberships. However, most 
of these
approaches focus on the identification of the most relevant concepts but do not provide concise, formal explanations of how such concepts are leveraged by the classifier to make predictions. In this paper, we propose a novel end-to-end differentiable approach enabling the extraction of logic explanations from neural networks using the formalism of First-Order Logic. The method relies on an entropy-based criterion which automatically identifies the most relevant concepts. We consider four different case studies to 
demonstrate that: (i) this entropy-based criterion 
enables
the distillation of concise logic explanations in safety-critical domains from clinical data to computer vision; (ii) the proposed approach outperforms state-of-the-art white-box models in terms of classification accuracy and matches black box performances.
\end{abstract}

\section{Introduction}

% intro to XAI
% \gc{Riprendere latex neurip 2021 con domande al fondo a cui rispondere.}

The lack of transparency in the decision process of some machine learning models, such as neural networks, limits their application in many safety-critical domains \citep{gdpr2017,goddard2017eu}. %,law10code}. 
For this reason, explainable artificial intelligence (XAI) research has focused either on \textit{explaining} black box decisions \citep{zilke2016deepred, ying2019gnnexplainer, ciravegna2020human, arrieta2020explainable} % ras2018explanation,
or on developing machine learning models \textit{interpretable by design} % \citep{breiman1984classification, 
\citep{schmidt2009distilling, letham2015interpretable, cranmer2019learning, molnar2020interpretable}. However, while interpretable models engender trust in their predictions \citep{doshi2017towards,doshi2018considerations,ahmad2018interpretable, rudin2021interpretable}, %samek2020toward, rudin2019stop
black box models, such as neural networks, are the ones that % models like neural networks generally
provide state-of-the-art task performances \citep{battaglia2018relational,devlin2018bert,dosovitskiy2020image,xie2020self}. Research to address this 
imbalance is % antithesis is urgently 
needed for the deployment of cutting-edge technologies.
% \citep{marcinkevivcs2020interpretability} -\textgreater minitour

\begin{figure*}[!ht]
    \centering
    \includegraphics[width=0.7\textwidth]{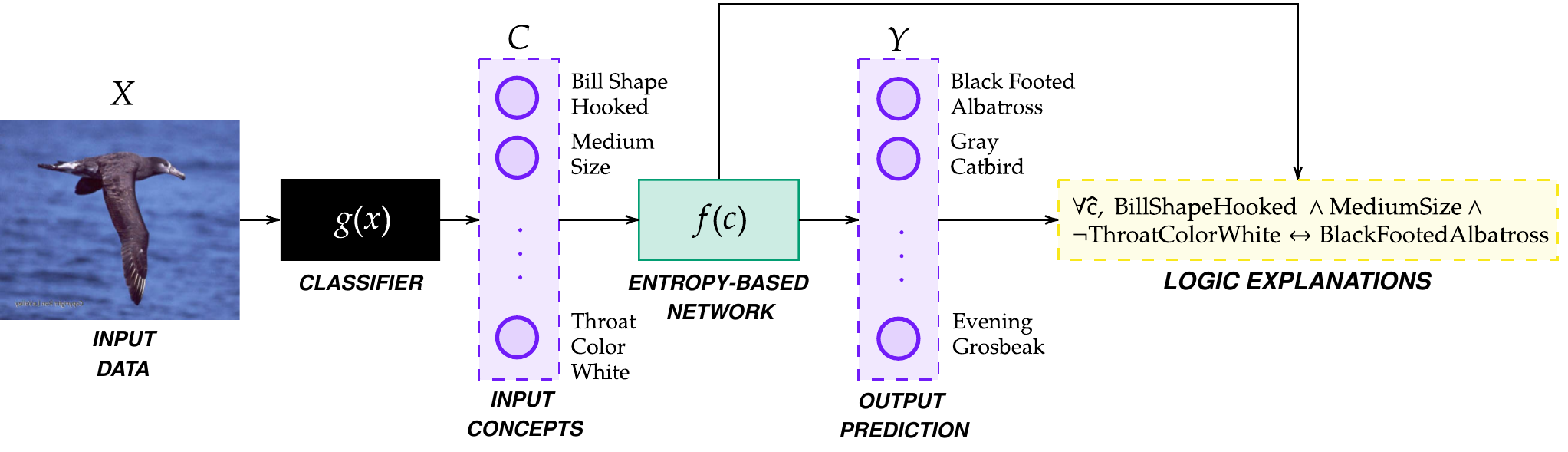}
    \caption{The proposed pipeline on one example from the CUB dataset. The neural network $f\colon C \mapsto Y$ maps concepts onto target classes and provide concise logic explanations (yellow -- arguments of predicates are dropped for simplicity) of its own decision process. When 
    % the features of the 
    the input data is non-interpretable (as pixels intensities), a classifier $g\colon X \mapsto C$ maps inputs to concepts.}
    \label{fig:abstract}
\end{figure*}

Most techniques \textit{explaining} black boxes focus on finding or ranking the most relevant features used by the model to make predictions \citep{simonyan2013deep, zeiler2014visualizing, ribeiro2016model, lundberg2017unified, selvaraju2017grad}. Such feature-scoring methods are very efficient and widely used, but they cannot explain how neural networks compose such features to make predictions \citep{kindermans2019reliability,kim2018interpretability,alvarez2018towards}.
In addition, a key issue of most \textit{explaining} methods is that explanations are given in terms of input features (e.g. pixel intensities) that do not correspond to high-level categories that humans can easily understand \citep{kim2018interpretability,su2019one}. To overcome this issue, \textit{concept-based} approaches have become increasingly popular as they provide explanations in terms of human-understandable categories (i.e. the \textit{concepts}) rather than raw features
\citep{kim2018interpretability,ghorbani2019towards,koh2020concept,chen2020concept}. % goyal2019explaining, yeh2020completeness
However, fewer approaches are able to explain how such concepts are leveraged by the classifier and even fewer provide concise explanations whose validity can be assessed quantitatively \citep{ribeiro2016model, guidotti2018local, das2020opportunities}. % zilke2016deepred, ciravegna2020constraint,

% interpretability by design?
% In other words, they cannot provide an explanation of the decision process of a neural networks.
% Further, Such feature ranking methods do not enable deeper human-machine interactions. Human experts can not make use of ranking of hundreds of features to understand the predictions of a network, nor the simple fact that some of them were important can be a sufficient form of explanation.

{\bf Contributions. } In this paper, we first propose an entropy-based % linear
 layer (Sec. \ref{sec:con_awa}) 
that enables % allowing 
the implementation of \textit{concept-based} neural networks, 
providing First-Order Logic explanations (Fig. \ref{fig:abstract}). 
The proposed approach is not just a post-hoc method, but an \textit{explainable by design} approach as it embeds additional constraints both in the architecture and in the learning process, to allow the emergence of simple logic explanations. This point of view is in contrast with post-hoc methods, which generally do not impose any constraint on classifiers: After the training is completed, the post-hoc method kicks in.
% We define our approach as \textit{explainable by design}, since the proposed architecture allows neural networks to automatically provide logic explanations of their predictions.
%supports the \textit{design} of more interpretable architectures enabling the straightforward generation of \textit{logic explanations}. 
Second, we describe how to interpret the predictions of the proposed neural model to distill logic explanations for individual observations
% , groups of samples, \fg{aggreg su cluster è stato rimosso dalla 3.3}
and for a whole target class (Sec. \ref{sec:fol}). We demonstrate how the proposed 
approach provides high-quality explanations according to six \textit{quantitative} metrics while matching black-box and outperforming state-of-the-art white-box models (Sec. \ref{sec:rel}) in terms of classification accuracy on four case studies (Sec. \ref{sec:experiments}).
Finally, we share an implementation of the entropy layer, with extensive documentation and all the experiments in the public repository: \url{https://github.com/pietrobarbiero/entropy-lens}.
% Finally, in the one and only benchmark with available ground-truth logic explanations, we observe how only the proposed approach provides explanations matching the ground-truth logic formulas.

\section{Background}
\label{sec:background}

Classification is the problem of identifying a set of categories an observation belongs to.
We indicate with $Y\subset \{0,1\}^r$ the space of binary encoded targets in a problem with $r$ categories.
Concept-based classifiers $f$ are a family of machine learning models predicting class memberships from 
the activation scores of $k$
human-understandable categories,
$f: C \mapsto Y$, 
where $C\subset[0,1]^k$ (see Fig. \ref{fig:abstract}). Concept-based classifiers improve human understanding as their input and output spaces consists of interpretable symbols. When observations are represented in terms of non-interpretable input features belonging to $X\subset\mathbb{R}^d$ (such as pixels intensities), a ``concept decoder'' $g$ is used to map the input into a concept-based space, $g: X \mapsto C$ (see Fig. \ref{fig:abstract}). Otherwise, they are simply rescaled from the unbounded space $\mathbb{R}^d$ 
into the unit interval $[0, 1]^k$, such that input features can be treated as logic predicates.

% Classification is the problem of identifying a set of categories $y \in Y\subset \{0,1\}^r$ an observation $x \in X\subset\mathbb{R}^d$ belongs to. A standard neural network is a black-box model $m: X \mapsto Y$ predicting for any sample $x\in X$ the corresponding class membership $\hat{y}\in Y$. In case the input features $x$ are not easily interpretable (low-level feature, image pixels) concept-based classifiers have been introduced to predict class memberships $Y$ from human-understandable categories  (a.k.a. \textit{concepts}) $C\subset[0,1]^k$: $f: C \mapsto Y$ to improve the understanding of black boxes and their decision process. Concepts can either correspond to the predictions of a classifier (i.e. $g: X \mapsto C$) \citep{koh2020concept} or simply to a re-scaling of the inputs space from the unbounded $\mathbb{R}^d$ to the unit interval $[0, 1]^k$ such that input features can be treated as logic predicates. Concept-based classifiers improve human understanding as their input and output spaces consists of interpretable symbols. 
% XAI methods such as LIME or SHAP can find the most relevant subset of concepts for the prediction of each category $Y^i$, but cannot explain how such concepts are combined by the classifier. In other words they cannot provide an explanation of the decision process.

In the recent literature, the most similar method related to the proposed approach is the $\psi$ network proposed by Ciravegna et al. \citep{ciravegna2020human, ciravegna2020constraint}, an end-to-end differentiable concept-based classifier \textit{explaining its own decision process}. The $\psi$ network leverages the intermediate symbolic layer whose output belongs to $C$ to distill First-Order Logic formulas, representing the learned map from $C$ to $Y$. The model consists of a sequence of fully connected layers with sigmoid activations only. An $L1$-regularization and a strong pruning strategy is applied to each layer of weights in order to allow the computation of logic formulas representing the activation of each node. 
% and, recursively, to generate concise explanations of the network predictions. 
Such constraints, however, limit the learning capacity of the network and impair the classification accuracy, making standard white-box models, such as decision trees, more attractive.

\section{Entropy-based Logic Explanations of Neural Networks} 
\label{sec:method}

\begin{figure}[t]
    \centering
    \includegraphics[width=0.9\columnwidth]{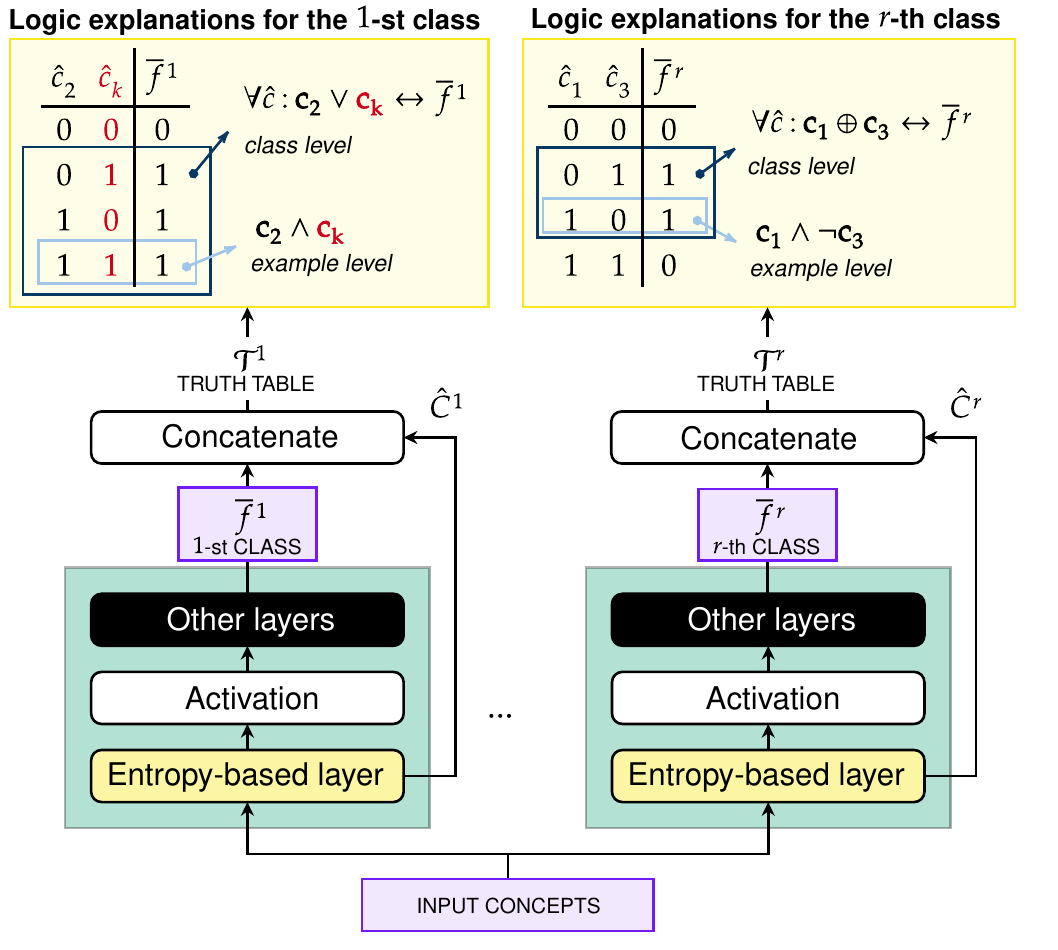}
    \caption{
    % The proposed % concept-based 
    % neural network learns the function $f:C\mapsto Y$.
    For each class $i$, the network leverages one ``head'' of the entropy-based linear layer (green) as first layer, and it provides: the class membership predictions $f^i$ and the truth table $\mathcal{T}^i$ (Eq. \ref{eq:truth-table}) to distill FOL explanations (yellow, top). 
    }
    \vskip -0.2 cm
    \label{fig:awareness}
\end{figure}

The key contribution of this paper is a novel linear layer enabling entropy-based logic explanations of neural networks (see Fig. \ref{fig:awareness} and Fig. \ref{fig:awareness2}). The layer input belongs to % is applied on 
the concept space $C$ and
the outcomes of the layer computations are: % produces: 
(i) the embeddings $h^i$ (as any linear layer), (ii) a truth table $\mathcal{T}^i$ explaining how the network leveraged concepts to make predictions for the $i$-th target class. 
Each class of the problem requires an independent entropy-based layer, as emphasized by the superscript $i$.
For ease of reading and without loss of generality, 
all the following descriptions % the following sections 
concern 
inference % the forward pass 
for 
a single % \textit{only one} 
observation (corresponding to the concept tuple $c \in C$) and 
a neural network $f^i$ % \textit{only one} network $f^i$ 
predicting the class memberships for the $i$-th class of the problem. %In practice, each classification objective requires its own entropy-based linear layer. 
For multi-class problems, multiple ``heads'' of this layer are instantiated, with one ``head'' per target class (see Sec. \ref{sec:experiments}), and the hidden layers of the class-specific networks could be eventually shared.

% Since inputs and outputs are vectors of $[0,1]$-values, any task function $f_i$ is almost interpretable-by-design and can be easily converted into a boolean function $\bar{f}_i$, e.g. by approximating to nearest crisp values, and in turn into a FOL formula $\varphi_i$ (see Sec. \ref{sec:fol}).
% However, to get readable formulas, any class prediction $f_i$ should only depend on a few concepts. CANNs, therefore, aim at providing a FOL explanation for any $f_i$ with respect to a (small) subset of the concepts in $C$.

\subsection{Entropy-based linear layer}
\label{sec:con_awa}
When humans compare a set of hypotheses outlining the same outcomes, they tend to have an implicit bias towards the simplest ones as outlined in philosophy \citep{soklakov2002occam,rathmanner2011philosophical},
% aristotlePosterior, %hoffmann1996ockham,
psychology \citep{miller1956magical,cowan2001magical}, and decision making \citep{simon1956rational,simon1957models,simon1979rational}.
% , information theory \citep{mackay2003information}, and natural sciences \citep{wiley2011phylogenetics,ma2014changing}. 
The proposed entropy-based approach encodes this inductive bias in an end-to-end differentiable model. The purpose of the entropy-based linear layer is to encourage the neural model to pick a limited subset of input concepts, allowing it to provide concise explanations of its predictions. The learnable parameters of the layer are the usual weight matrix $W$ and bias vector $b$. In the following, the forward pass is described by the operations going from Eq. \ref{eq:gamma} to Eq. \ref{eq:forward}
while % and 
the generation of the truth tables from which explanations are extracted 
is formalized by % in 
Eq. \ref{eq:sparse} and Eq. \ref{eq:truth-table}.

\begin{figure}[t]
    \centering
    \includegraphics[width=0.65\columnwidth]{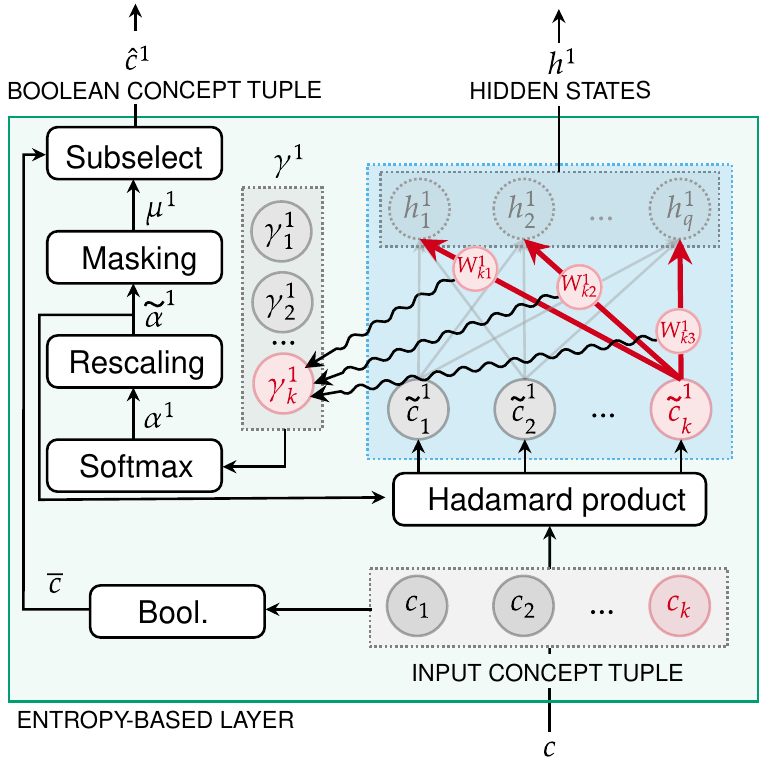}
    \caption{A detailed view on one ``head'' of
    the entropy-based linear layer for the $1$-st class, emphasizing the role of the $k$-th input concept as example: (i) the scalar $\gamma_k^1$ (Eq.~\ref{eq:gamma}) is computed from the 
    set of weights connecting the $k$-th input concept 
    to the output neurons of the entropy-based layer;
    % $weight vector $W_k^1$, the weights associated to the $k$-th concept (Eq. \ref{eq:gamma}); 
    (ii) the relative importance of each concept is summarized by the categorical distribution $\alpha^1$ (Eq. \ref{eq:alpha}); (iii) rescaled relevance scores $\tilde{\alpha}^1$ drop irrelevant input concepts out (Eq. \ref{eq:drop}); (iv) hidden states $h^1$ (Eq. \ref{eq:forward}) and Boolean-like concepts $\hat{c}^1$ (Eq. \ref{eq:sparse}) are provided as outputs of the entropy-based layer.}
    \label{fig:awareness2}
\end{figure}

The relevance of each input concept can be summarized in a first approximation by a measure that depends on the values of the weights connecting such concept to the upper network. In the case of network $f^i$ (i.e. predicting the $i$-th class) and of the $j$-th input concept, we indicate with $W_j^i$ the vector of weights departing from the $j$-th input (see Fig. \ref{fig:awareness2}), and we introduce
%For the network $f^i$ (i.e. predicting the $i$-th class) the relevance of the $j$-th input concept is summarized in first approximation by the vector $W_j^i$ representing the weights connecting the $j$-th concept with the first layer of hidden neurons (see Fig \ref{fig:awareness}):
\begin{equation} \label{eq:gamma}
    \gamma^i_j = ||W^i_j||_1\ .
\end{equation}
The higher $\gamma^i_j$, the higher the relevance of the concept $j$ for the network $f^i$. In the limit case ($\gamma_j^i \rightarrow 0$) the model $f^i$ drops the $j$-th concept out.
% Notice that since the vector $\gamma^i$ is computed for each class, hidden network layers are not shared among network outputs but are independent as shown in Fig \ref{fig:awareness}.
To select only few relevant concepts for each target class, concepts are set up to compete against each other. To this aim, the relative importance of each concept to the $i$-th class is summarized in the categorical distribution 
$\alpha^{i}$, composed of coefficients
$\alpha^i_j \in [0,1]$ (with $\sum_j \alpha_j^i = 1$), modeled by the softmax function:
\begin{equation} \label{eq:alpha}
    \alpha^i_j = \frac{e^{\gamma^i_j/\tau}}{\sum_{l=1}^k e^{\gamma^i_l/\tau}}
\end{equation}
where $\tau \in \mathbb{R}^+$ is a user-defined temperature parameter to tune
% the intrinsic tendency of 
the softmax function. \pb{For a given set of $\gamma^i_j$, when using} high temperature values ($\tau \rightarrow \infty$) all concepts have nearly the same relevance. For low temperatures values ($\tau \rightarrow 0$), the probability of the most relevant concept tends to $\alpha_j^i\approx 1$, while it becomes $\alpha_k^i\approx 0, \ k \neq j$, for all other concepts. For further details on the impact of $\tau$ on the model predictions and explanations, see Appendix \ref{appendix:hyperparam_explanations}.
As the probability distribution $\alpha^i$ highlights the most relevant concepts, this information is directly fed back to the input, weighting concepts by the estimated importance. To avoid numerical cancellation due to values in $\alpha^i$ close to zero, especially when the input dimensionality is large,
we replace $\alpha^i$ with its normalized instance $\tilde{\alpha}^i$, still  in $[0,1]^k$,
and each input sample % represented by the concept tuple ù
$c \in C$ is modulated by this %(normalized) 
estimated importance, % weighted by the re-normalized vector $\hat{\alpha}^i \in [0,1]^k$:w
\begin{equation} \label{eq:drop}
    \tilde{c}^i = c \odot \tilde{\alpha}^i \qquad\qquad \text{with} \qquad \tilde{\alpha}_j^i = \frac{\alpha_j^i}{\max_u \alpha_u^i},
\end{equation}
where $\odot$ denotes the Hadamard (element-wise) product.
The highest value in $\tilde{\alpha}^i$ is always $1$ (i.e. $\max_j \tilde{\alpha}_j^i = 1$) and it corresponds to the most relevant concept. 
% All the other concepts are weighted by an $\hat{\alpha}_j^i \leq 1$. 
The embeddings $h^i$ are computed as in any linear layer by means of the affine transformation:
\begin{equation} \label{eq:forward}
    h^i = W^i \tilde{c}^i + b^i.
\end{equation}
Whenever $\tilde{\alpha}_j^i \rightarrow 0$, the input $\tilde{c}_j^i \rightarrow 0$.
%In the limit, as $\hat{\alpha}_j^i \rightarrow 0$, the input $\tilde{c}_j \rightarrow 0$. 
This means that the corresponding concept tends to be dropped out and the network $f^i$ will learn to predict the $i$-th class without 
relying on % using 
the $j$-th concept. 

In order to % To 
get logic explanations, the proposed linear layer generates the truth table $\mathcal{T}^i$ formally representing the behaviour of the neural network 
in \pb{terms} of Boolean-like representations of the input concepts. % for each category employed as classification objective. 
In detail, we indicate with $\bar{c}$ the Boolean interpretation of the input tuple $c \in C$, while $\mu^i \in \{0,1\}^k$ is the binary mask associated to $\tilde{\alpha}^i$.
To encode the inductive human bias towards simple explanations \citep{miller1956magical,cowan2001magical,ma2014changing}, the 
mask % vector
% $\mu^i \in [0,1]^k$ 
$\mu^i$
% is computed and 
is used to generate the 
binary % one-hot 
concept tuple $\hat{c}^i$, 
dropping % masking 
the least relevant concepts out of $c$,
\begin{equation}\label{eq:sparse}
    \hat{c}^i = \xi(\bar{c}, \mu^i)  \quad \text{with} \quad
    \mu^i = \mathbb{I}_{\tilde{\alpha}^i \geq \pb{\epsilon}} \quad \text{and} \quad \bar{c} = \mathbb{I}_{c \geq \pb{\epsilon}},
\end{equation}
where $\mathbb{I}_{z \geq \pb{\epsilon}}$ denotes the indicator function that is $1$ for all the components of vector $z$ being $\geq \pb{\epsilon}$ and $0$ otherwise (considering the unbiased case, we set $\pb{\epsilon}=0.5$).
The function $\xi$ returns the vector with the components of $\bar{c}$ that correspond to $1$'s in $\mu^i$ (i.e. it sub-selects the data in $\bar{c}$).
%while $\xi$ sub-selects the concepts for which $\mu^i_j=1$.
As a results, $\hat{c}^i$ belongs to a space $\hat{C}^i$ of $m_i$ Boolean features, with $m_i \textless k$ due to the effects of the subselection procedure.
%Given a dataset $\mathcal{D}=(\mathcal{C},\mathcal{Y})$, sparse concept representations $\hat{c}^i$ are obtained for each observation from Eq. \ref{eq:sparse} and stacked together in the sparse matrix $\hat{\mathcal{C}}^i$.
%= [\hat{c}^i(1), ...., \hat{c}^i(m)]$. 
%This matrix is concatenated with the Boolean\sm{-interpreted} model predictions $\bar{f}^i = \mathbb{I}_{f^i \geq 0.5}$ to obtain the matrix $\mathcal{T}^i$ corresponding to the sparse truth table used to generate logic explanations (see Sec. \ref{sec:fol}):

The truth table $\mathcal{T}^i$ is a particular way of representing the behaviour of network $f^i$ based on the outcomes of
 processing multiple input samples collected in a generic dataset $\mathcal{C}$.
 As the truth table involves Boolean data, we denote with 
$\hat{\mathcal{C}}^i$ the set with the Boolean-like representations of the samples in $\mathcal{C}$ computed by $\xi$, Eq.~\ref{eq:sparse}.
We also introduce $\bar{f}^i(c)$ as the Boolean-like representation of the network output, $\bar{f}^i(c)=\mathbb{I}_{f^i(c)\geq \pb{\epsilon}}$.
%old
% From an operational perspective, the contents $\mathbf{T}^i$ of the truth table $\mathcal{T}^i$ are obtained by stacking data of $\hat{\mathcal{C}}^i$ into a 2D matrix $\hat{\mathbf{C}}^i$ (row-wise), and concatenating the result with the column vector $\bar{\mathbf{f}}^i$ whose elements are $\bar{f}^i(c)$, $c\in \mathcal{C}$, that we summarize as
% \begin{equation} \label{eq:truth-table}
%     \mathbf{T}^i = \Big( \hat{\mathbf{C}}^i \ \Big|\Big| \ \bar{\mathbf{f}}^i \Big).
% \end{equation}
The truth table $\mathcal{T}^i$ is obtained by stacking data of $\hat{\mathcal{C}}^i$ into a 2D matrix $\hat{\mathbf{C}}^i$ (row-wise), and concatenating the result with the column vector $\bar{\mathbf{f}}^i$ whose elements are $\bar{f}^i(c)$, $c\in \mathcal{C}$, that we summarize as
\begin{equation} \label{eq:truth-table}
    \mathcal{T}^i = \Big( \hat{\mathbf{C}}^i \ \Big|\Big| \ \bar{\mathbf{f}}^i \Big).
\end{equation}
To be precise, any $\mathcal{T}^i$ is more like an empirical truth table than a classic one corresponding to an $n$-ary boolean function, indeed $\mathcal{T}^i$ can have repeated rows and missing Boolean tuple entries. However, $\mathcal{T}^i$ can be used to generate logic explanations in the same way, as we will explain in Sec. \ref{sec:fol}.
% The purpose of the probability distribution $\alpha$ is threefold: (i) it models the relative importance of concepts enabling the inspection of the most relevant ones for each classification task; (ii) it 
% \begin{equation}
%     \hat{\alpha}^i = \frac{\alpha^i}{\max \alpha^i}
% \end{equation}
% At training epoch $t$, therefore, the weighted concepts $\tilde{C}^i$ will be the input of the neural network $\hat{y}^i = f^i \big( \tilde{C}^i \big)$. 
% \fg{Ma il concept awareness non fa già parte della f?}
% Concept awareness scores close to zero will drop out the least relevant input concepts allowing for simpler logic-based explanations of the neural network's decisions, as explained in Sec. \ref{sec:fol}. 
% More precisely, only input concepts which satisfy the following condition are considered:
% \begin{equation}
%     E_i =  \left\{\langle c_j \rangle  \mid \frac{\alpha^i_j}{\max_j{\alpha^i}} \textgreater 0.5 \right\}, 
% \end{equation}
% where $E_i$ denotes the set of input concepts employed to explain the output $f_i$.
% \subsubsection{Multi-task awareness}

% \begin{figure}[h]
%     \centering
%     \includegraphics[width=0.6\textwidth]{figs/logic_layers.pdf}
%     \caption{Multi-Head awareness consists of several awareness layers running in parallel. GC: CAMBIARE NOME DEL LAYER NELLA FIGURA.}
%     \label{fig:multi_head}
% \end{figure}

% \subsubsection{Multi-Head awareness}

\subsection{Loss function}
The entropy of the probability distribution $\alpha^i$ (Eq. \ref{eq:alpha}),
\begin{equation}
    \mathcal{H}(\alpha^i) = - \sum_{j=1}^k \alpha^i_j \log \alpha^i_j
    \label{eq:ent}
\end{equation}
is minimized when a single $\alpha^i_j$ is one, thus representing the extreme case in which only one concept matters, while it is maximum when all concepts are equally important. When $\mathcal{H}$ is jointly minimized with the usual loss function for supervised learning $L(f,y)$ (being $y$ the target labels--we used the cross-entropy in our experiments), it allows the model to find a trade off between fitting quality and a parsimonious activation of the concepts, 
allowing each network $f^i$ to predict $i$-th class memberships using few relevant concepts only.
%The minimum of this function corresponds to a one-hot encoded configuration of $\alpha$. When $\alpha_j^i=0$ the $j$-th concept is not taken into consideration when predicting the $i$-th class, as described in Eq. \ref{eq:drop}.
%A cross entropy loss $L(f,y)$ is also employed on the available labels $y$ for standard supervised training.
%\sm{We define with $L(f,y)$ the cross-entropy loss for supervised data, being $y$ the target labels}.
Overall, the loss function to train the network $f$ is defined as,
% The overall loss function to train network $f$ is defined by summing Eq. \ref{eq:ent} with the cross entropy loss $L$ on the available labels $y$ for $f$ used for standard supervised training:
\begin{equation}
    \mathcal{L}(f,y,\alpha_1,\ldots,\alpha_r) = L(f,y) + \lambda \sum_{i=1}^r\mathcal{H}(\alpha^i),
    \label{eq:loss}
\end{equation}
where $\lambda \textgreater 0$ is the hyperparameter used to balance the relative importance of low-entropy solutions in the loss function. Higher values of $\lambda$ lead to sparser configuration of $\alpha$, constraining the network to focus on a smaller set of concepts for each classification task (and vice versa), thus encoding the inductive human bias towards simple explanations \citep{miller1956magical,cowan2001magical,ma2014changing}. For further details on the impact of $\lambda$ on the model predictions and explanations, see Appendix \ref{appendix:hyperparam_explanations}.
It may be pointed out that a similar regularization effect could be achieved by simply minimizing the $L_1$ norm over $\gamma^i$. However, as we observed in \ref{appendix:entropy_vs_L1}, the $L_1$ loss does not sufficiently penalize the concept scores for those features which are uncorrelated with the predicted category. The Entropy loss, instead, correctly shrink to zero concept scores associated to uncorrelated features while the other remains close to one.

\subsection{First-order logic explanations}
\label{sec:fol}
%\fg{WIP: todo invece delle stringhe "c_j" dire che i bold sono predicati sulle tuple di concetti in un dominio.. così nella regola logica phi_S(c) è un grounding di phi_S}

Any Boolean function can be converted into a logic formula in Disjunctive Normal Form (DNF) by means of its truth-table \citep{mendelson2009introduction}. 
% We indicate with $\hat{f}^i$ the Boolean function represented by the truth table $\mathcal{T}^i$, $\hat{f}^i: \hat{C}^i \mapsto Y^i$, being $Y^i$ the $i$-th component of $Y$.
%\sm{Following the notation defined at the end of Sec.~\ref{sec:con_awa}}, any $\bar{f}^i$ is a Boolean function over the set $\hat{C}^i$. %We denote with $\varphi_i$ the logic formula corresponding to the truth-table $\mathcal{T}^i$ of $\bar{f}^i$. 
Converting a truth table into a DNF formula provides an effective mechanism to extract logic rules of increasing complexity from individual observations
% , for cluster of samples,
to a whole class of samples. 
% In order to expliciteply write down the syntactic logic formula corresponding to any boolean function, we will use text strings in quotation marks corresponding to both concept and task symbols.
% The following steps are repeated for any task function $f^i$.
% [TODO] In the following, with a little abuse of notation, we will denote by $\bar{c}_j,\neg\bar{c}_j$ and $\bar{f}^i,\neg\bar{f}^i$ both the Boolean values and the human-understandable concept and task name and their negated, respectively, for every $j,i$. 
The following rule extraction mechanism is applied to any empirical truth table $\mathcal{T}^i$ for each task $i$.
% \begin{figure}[h]
%     \centering
%     \includegraphics[width=1\textwidth]{figs/truth-table.png}
%     \caption{Caption}
%     \label{fig:tt2form}
% \end{figure}

% %\vspace{-3mm}
\paragraph{FOL extraction.}
Each row of the truth table $\mathcal{T}^i$ can be partitioned into two parts that are a tuple of binary concept activations, $\hat{q}\in \hat{C}^i$, and the outcome of $\bar{f}^i(\hat{q}) \in \{0, 1\}$. 
An \textit{example-level} logic formula, consisting in a single minterm, can be trivially extracted from each row for which $\bar{f}^i(\hat{q})=1$, by simply connecting with the logic AND ($\wedge$) the true concepts and negated instances of the false ones. 
The logic formula becomes human understandable whenever concepts appearing in such a formula are replaced with human-interpretable strings that represent their name (similar consideration holds for $\bar{f}^i$, in what follows). For example, the following logic formula $\varphi^i_t$,
\begin{equation}
    \varphi^i_{t} = \textbf{c}_1\wedge\ \neg \textbf{c}_2 \wedge \ldots\wedge\textbf{c}_{m_i},
    \label{eq:locexp}
\end{equation}
is the formula extracted from the $t$-th row of the table where, in the considered example, only the second concept is false, being $\textbf{c}_z$ the name of the $z$-th concept.
Example-level formulas can be aggregated with the logic OR ($\vee$) to provide a \textit{class-level} formula, 
\begin{equation}
    \displaystyle\bigvee_{t \in S_i}\varphi^i_t, %= \displaystyle\bigvee_{\hat{c} \in S_i}\textbf{c}_1\wedge\ldots\wedge\textbf{c}_{m_i}
\label{eq:agg_exp}
\end{equation}
being $S_i$ the set of rows
 indices of $\mathcal{T}^i$ for which $\bar{f}^i(\hat{q}) = 1$, i.e. it is the support of $\bar{f}^i$.
%Once $\varphi^i$ is built, given a concept tuple $\bar{c} \in \hat{C}^i$, we may measure the satisfiability of $\varphi^i$ on $\bar{c}$ by the truth-value of $\varphi^i(\bar{c})$ obtained by replacing boolean values with concept names in Eq. \ref{eq:agg_exp}. 
%By construction the support of the class-level rule and of $\hat{f}^i$ coincide.
We define with $\phi^i(\hat{c})$ the function that holds true whenever Eq.~\ref{eq:agg_exp}, evaluated on a given Boolean tuple $\hat{c}$, is true.
Due to the aforementioned definition of support, we get the following class-level First-Order Logic (FOL) explanation for all the concept tuples,
\begin{equation}
\forall \hat{c} \in \hat{C}^i:\ \phi^i(\hat{c})\leftrightarrow\bar{f}^i(\hat{c}).
\label{eq:FOL_C}
\end{equation}
We note that in case of non-concept-like input features, we may still derive the FOL formula through the ``concept decoder'' function $g$ (see Sec. \ref{sec:background}),
\begin{equation}
\forall x \in X:\ \phi^i\left(\xi(\overline{g(x)},\mu^i)\right)\leftrightarrow\bar{f}^i\left(\xi(\overline{g(x)},\mu^i)\right)
\end{equation}
An example of the above scheme for both example and class-level explanations is depicted on top-right of Fig. \ref{fig:awareness}.

%\vspace{-3mm}
\paragraph{Remarks.} The aggregation of many example-level explanations may increase the length and the complexity of the FOL formula being extracted for a whole class. However, existing techniques as the Quine–McCluskey algorithm can be used to get compact and simplified equivalent FOL expressions \citep{mccoll1878calculus,quine1952problem,mccluskey1956minimization}. For instance, the explanation (\textit{person} $\wedge$ \textit{nose}) $\vee$ ($\neg$\textit{person} $\wedge$ \textit{nose}) can be formally simplified in \textit{nose}.
%\fg{mettere o omettere questo paragrafo?}
%\paragraph{Remarks.}
Moreover, the Boolean interpretation of concept tuples may generate colliding representations for different samples. For instance, the Boolean representation of the two samples $\{ (0.1, 0.7), (0.2, 0.9) \}$ is the tuple $\bar{c} = (0, 1)$ for both of them. This means that their example-level explanations match as well. %It is worth to notice that formally Eq. \ref{eq:FOL_C} relies on the fact that there are no two different concept tuples $c,d$ with $\bar{c}=\bar{d}$ such that $f^i(c)\neq f^i(d)$ for some $i$. 
However, a concept can be eventually split into multiple finer grain concepts to avoid collisions. Finally, we mention that the number of samples for which any example-level formula holds (i.e. the support of the formula) is used as a measure of the explanation importance. In practice, example-level formulas are ranked by support and iteratively aggregated to extract class-level explanations, until the aggregation improves the accuracy of the explanation over a validation set.

\vspace{-1mm}
\section{Related work}
\label{sec:rel}
% \begin{itemize}
% \item concept-bottleneck: noi si da più interpretabilità, perchè loro sono limitati alla logistic regression (che non è in grado di rappresentare dipendenze tra concetti)
%     \item psi networks, mu networks, relu networks: are all special cases (e noi si risultati migliori e no-pruning)
%     \item CBM, TCAV, NYSM, concept whitening: interesting to get insights on the model behaviour, but not to make decisions!
%     \item decision trees, BRL: much better performance, built-in in the architecture
%     \item multi-task feature-selection, feature importance, $L_1$ regularization 
% \end{itemize}
% In the last few years, the demand for human-comprehensible models has significantly increased in safety-critical and data-sensible contexts. 
In order to provide explanations for a given black-box model, most methods focus on identifying or scoring the most relevant input features \citep{simonyan2013deep, zeiler2014visualizing, ribeiro2016model, ribeiro2016should, lundberg2017unified, selvaraju2017grad}. Feature scores are usually computed sample by sample (i.e. providing \textit{local explanations}) analyzing the activation patterns in the hidden layers of neural networks \citep{simonyan2013deep, zeiler2014visualizing, selvaraju2017grad} or by 
% considering the network as a black box
following a model-agnostic approach \citep{ribeiro2016should, lundberg2017unified}. 
To enhance human understanding of feature scoring methods, concept-based approaches have been effectively employed for identifying common activations patterns in the last nodes of neural networks corresponding to human categories \citep{kim2018tcav, kazhdan2020now} or constraining the network to learn such concepts \citep{chen2020concept, koh2020concept}.
Either way, feature-scoring methods are not able to explain \textit{how} neural networks compose features to make predictions \citep{kindermans2019reliability,kim2018interpretability,alvarez2018towards} and only a few of these approaches have been efficiently extended to provide explanations for a whole class (i.e. providing \textit{global explanations}) \citep{simonyan2013deep, ribeiro2016should}. 
% of scarce utility in decision support cases, where a comprehensible language is of crucial importance to understand correlations and the causality in the decision process of the classifier.
% From a different prospective, some recent approaches attempted to identify common activations patterns in the last nodes of a neural network which could be associated to symbolic concepts \citep{kim2018tcav, kazhdan2020now}, or to directly force the network to extract such concepts \citep{chen2020concept, koh2020concept}. Also in this case, even though these methods may help in understanding the model behaviour, they can not be employed to support human decisions. 
% Differently, rule-based explanations can be employed in such a scenario since they usually rely on a formal language, such as FOL. 
By contrast, a variety of rule-based approaches have been proposed to provide concept-based explanations.
Logic rules are used to explain how black boxes predict class memberships for indivudal samples \citep{guidotti2018local, ribeiro2018anchors}, or for a whole class \citep{sato2001rule, zilke2016deepred, ciravegna2020human, ciravegna2020constraint}. 
Distilling explanations from an existing model, however, is not the only way to achieve explainability. Historically, standard machine-learning such as Logistic Regression \citep{mckelvey1975statistical}, Generalized Additive Models \citep{hastie1987generalized, lou2012intelligible, caruana2015intelligible} Decision Trees \citep{breiman1984classification, quinlan1986induction, quinlan2014c4} and Decision Lists \citep{rivest1987learning, letham2015interpretable, angelino2018learning} were devised to be intrinsically interpretable. However, most of them struggle in solving complex classification problems. Logistic Regression, for instance, in its vanilla definition, can only recognize linear patterns, e.g. it cannot to solve the XOR problem \citep{minsky2017perceptrons}. Further, only Decision Trees and Decision Lists provide explanations in the from of logic rules. Considering decision trees, each path may be seen as a human comprehensible decision rule when the height of the tree is reasonably contained. 
Another family of concept-based XAI methods is represented by rule-mining algorithms which became popular at the end of the last century \citep{holte1993very, cohen1995fast}. Recent research has led to powerful rule-mining approaches as Bayesian Rule Lists (BRL) \citep{letham2015interpretable}, where a set of rules is ``pre-mined'' using the frequent-pattern tree mining algorithm \citep{han2000mining} and then the best rule set is identified with Bayesian statistics.
In this paper, the proposed approach is compared with methods providing logic-based, global explanations.
% for which an efficient implementation is available. 
In particular, we selected one representative approach from different families of methods: Decision Trees\footnote{\url{https://scikit-learn.org/stable/modules/tree}.} (white-box machine learning), BRL\footnote{\url{https://github.com/tmadl/sklearn-expertsys.}} (rule mining) and $\psi$ Networks\footnote{\url{https://github.com/pietrobarbiero/logic_explainer_networks.}}
% \footnote{In \citep{ciravegna2020human}, $\psi$ networks were devised to provide explanations of existing models; in the following, we show that they can directly solve classification problems.} 
(explainable neural models).
% \textcolor{red}{TO CHECK: Feature-scoring methods \citep{ribeiro2016model,selvaraju2017grad,alvarez2018towards,goyal2019counterfactual,kanehira2019learning} are not taken into consideration as no meaningful quantitative metrics exists to the best of our knowledge to evaluate these kind of explanations.}

\vspace{-1mm}
\section{Experiments}
\label{sec:experiments}

\begin{figure}[t]
    \centering
    \includegraphics[width=0.8\columnwidth]{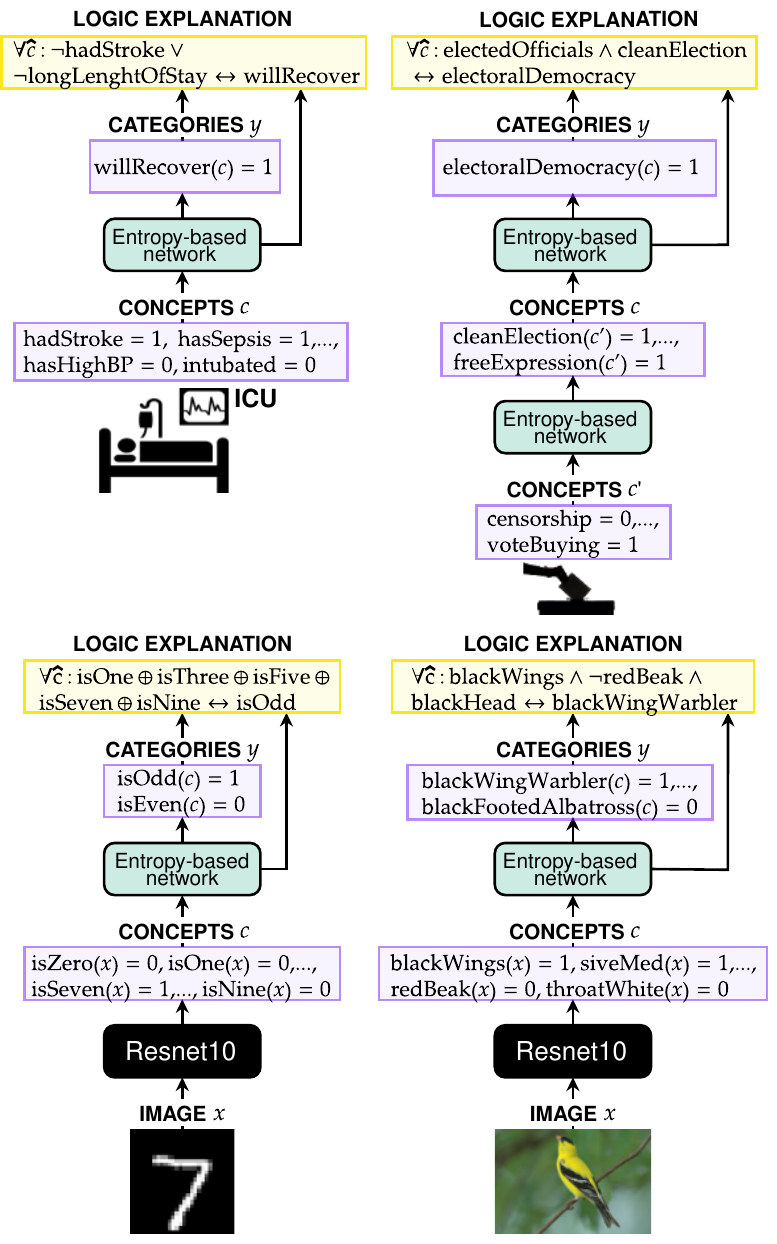}
    \caption{The four case studies show how the proposed Entropy-based networks (green) provide concise logic explanations (yellow) %---we dropped the arguments in the logic predicates, for simplicity) 
    of their own decision process in different real-world contexts. When input features are non-interpretable, as pixel intensities, a ``concept decoder'' (ResNet10) %is employed to 
    maps images into concepts. Entropy-based networks then map concepts into target classes.}
    % and provide the truth table to generate logic explanations.
    \label{fig:experiments}
    \vspace{-2 mm}
\end{figure}

The quality of the explanations and the classification performance of the proposed approach are quantitatively assessed and compared to state-of-the-art white-box models. A visual sketch of each classification problem (described in detail in Sec. \ref{sec:exp_details}) %together with all the experimental details)
and a selection of the logic formulas found by the proposed approach is reported in Fig. \ref{fig:experiments}. Six quantitative metrics are defined and used to compare the proposed approach with state-of-the-art methods. % A brief description of each classification problem, the corresponding dataset and all the related experimental details are exposed in Sec. \ref{sec:exp_details}. 
Sec. \ref{sec:results} summarizes the main findings. Further details concerning the experiments are reported in the supplemental material \ref{appendix}.
% A python package and a freely available GitHub repository implementing the proposed approach will be made public upon  paper acceptance. In appendix \ref{appendix:software} a snippet of the code extracted from the library is reported.

% \begin{table}[ht]
% \centering
% \caption{Experiments.}
% \label{tab:datasets}
% % \resizebox{\textwidth}{!}{
% \begin{tabular}{lll}
% \toprule
% \textbf{Dataset} & \textbf{Description} & \textbf{Pipeline}\\
% \midrule

% MIMIC-II  & \small{Predict patient survival from clinical data}      & End-to-End Model\\
% MNIST E/O      & \small{Predict parity from digit images}             & Concept-Bottleneck Model\\
% CUB  & \small{Predict bird species from bird images}            & Concept-Bottleneck Model\\
% V-Dem  & \small{Predict electoral democracy from social indexes}  & Cascading Models\\
% \bottomrule
% \end{tabular}
% % }
% \end{table}

\begin{table*}[t]
\small
\centering
% \caption{Classification accuracy (\%) of the compared models.}
\caption{Classification accuracy (\%). Left group, the compared white-box models. Right group, two black box models. We indicate in bold the best model in each group, with a star the best model overall.}
\label{tab:model-accuracy}
\vspace{-2mm}
% \footnotesize
\begin{tabular}{l|llll|ll}
\toprule
{} &         Entropy net   &        Tree &              BRL &                $\psi$ net & \pb{Neural Network}  & Random Forest\\
\midrule
\textbf{MIMIC-II     }  &  ${\bf 79.05^\star \pm 1.35}$  &  $77.53 \pm 1.45$ &  $76.40 \pm 1.22$ &  $77.19 \pm 1.64$ & $\pb{77.81 \pm 2.45}$ &${\bf 78.88 \pm 2.25}$\\
\textbf{V-Dem         } &  ${\bf 94.51 \pm 0.48}$  &  $85.61 \pm 0.57$ &  $91.23 \pm 0.75$ &  $89.77 \pm 2.07$ & $\pb{\bf 94.53^\star \pm 1.17}$ &$93.08 \pm 0.44$ \\
\textbf{MNIST}          &  ${\bf 99.81 \pm 0.02}$ &  $99.75 \pm 0.01$ &  $99.80 \pm 0.02$ &  $99.79 \pm 0.03$ & %\pb{$99.81 \pm 0.08$} \\
${ 99.72 \pm 0.03}$ & ${\bf 99.96^\star \pm 0.01}$ \\
\textbf{CUB          }  &  ${\bf 92.95 \pm 0.20}$  &  $81.62 \pm 1.17$ &  $90.79 \pm 0.34$ &  $91.92 \pm 0.27$ & %\pb{$93.32 \pm 0.35$}
${\bf {93.10}^{\star} \pm 0.51}$ & $91.88 \pm 0.36$  \\
\bottomrule
\end{tabular}
\vspace{-1.5 mm}
\end{table*}

\subsection{Classification tasks and datasets}
\label{sec:exp_details}

% For the experimental analysis, 
Four classification problems ranging from computer vision to medicine are considered. Computer vision datasets (e.g. CUB) are annotated with low-level concepts (e.g. bird attributes) used to train concept bottleneck pipelines \citep{koh2020concept}. 
In the other datasets, the input data is rescaled into a categorical space ($\mathbb{R}^k \rightarrow C$) suitable for concept-based networks. 
Please notice that this preprocessing step is performed for all white-box models considered in the experiments for a fair comparison. Further descriptions of each dataset and links to all sources are reported in Appendix \ref{appendix:dataset}.\\
%\vspace{-3mm}
\textbf{Will we recover from ICU? (MIMIC-II).}
The Multiparameter Intelligent Monitoring in Intensive Care II (MIMIC-II, \citep{saeed2011multiparameter,goldberger2000physiobank}) is a public-access intensive care unit (ICU) database consisting of 32,536 subjects (with 40,426 ICU admissions) admitted to different ICUs.
%medical, surgical, cardiovascular and neonatal ICUs, surgical recovery units, and coronary care units ICUs at a single tertiary care hospital. 
% The dataset contains detailed descriptions of a variety of clinical data classes: general,
%(e.g. patient demographics, hospital admission and discharge dates, hospital death dates, etc), 
% physiological, % (e.g.	nurse-verified vital signs, blood pressure, heart rate, etc), 
% results of clinical laboratory tests, % (e.g. hematology, blood chemistries, urinalysis, microbiology, etc), 
% records of medications, fluid balance, %(e.g. solutions, urine, estimated blood loss, etc), 
% and text reports of imaging studies (e.g. x-ray, CT, MRI, etc). In our experiments, we removed non-anonymous information, text-based features, time series inputs, and observations with missing data.
%discretize some of the continuous features into a set of categories in order to get a one-hot encoded representation. For instance, 
% We discretize continuous features
% , like blood pressure (BP), 
% into one-hot encoded categories.
% (e.g. very low BP, low BP, normal BP, high BP, and very high BP). 
% After such preprocessing step, we obtained an input space $C$ composed of $k=90$ key features. 
The task consists in 
% training a classifier function to 
identifying recovering %($Y=1$) 
or dying %($Y=0$) 
patients after ICU admission. 
An end-to-end classifier $f: C \rightarrow Y$ carries out the classification task.\\
%\vspace{-3mm}
\textbf{What kind of democracy are we living in? (V-Dem).}
Varieties of Democracy  (V-Dem, \citep{pemstein2018v,coppedge2021v}) dataset contains a collection of indicators of latent regime characteristics over 202 countries from 1789 to 2020. 
The database include $k_1 = 483$ low-level indicators %(e.g. media bias, party ban, high-court independence,  etc.), 
$k_2 = 82$ mid-level indices.
%(e.g. freedom of expression, freedom of association, equality before the law, etc), and 5 high-level indices of democracy principles (i.e. electoral, liberal, participatory, deliberative, and egalitarian). 
The task consists in identifying electoral democracies %($Y=1$)
from non-electoral ones. %($Y=0$). 
We indicate with $C_1$, $C_2$ the spaces associated to the activations of the %aforementioned 
two levels of concepts. Classifiers $f_1$ and $f_2$ are trained to learn the map $C_1 \rightarrow C_2 \rightarrow Y$. Explanations are given for classifier $f_2$ in terms of concepts $c_2 \in C_2$.\\
% \paragraph{MNIST Even/Odd}
%\vspace{-3mm}
\textbf{What does parity mean? (MNIST Even/Odd).}
The Modified National Institute of Standards and Technology database (MNIST, \citep{lecun1998mnist}) contains a large collection of images representing handwritten digits. 
% The input space $X \subset \mathbb{R}^{28\times 28}$ is composed of 28x28 pixel images while the concept space $C$ with $k=10$ is represented by the label indicator for digits from $0$ to $9$. 
% However, 
The task we consider here is slightly different from the common digit-classification. Assuming $Y \subset \{ 0,1 \}^2$, we are interested in determining if a digit is either odd or even, and explaining the assignment to one of these classes in terms of the digit labels (concepts in $C$). 
The mapping $X \rightarrow C$ is provided by a ResNet10 classifier $g$ \citep{he2016deep} trained from scratch.
while the classifier $f$ learn both the final mapping and the explanation as a function $C \rightarrow Y$.\\
%\vspace{-3mm}
\textbf{What kind of bird is that? (CUB).}
The Caltech-UCSD Birds-200-2011 dataset (CUB, \citep{wah2011caltech}) is a fine-grained classification dataset. It includes 11,788 images representing $r = 200$ ($Y = \{0,1\}^{200}$) different bird species. 312 binary attributes (concepts in $C$) describe visual characteristics (color, pattern, shape) of particular parts (beak, wings, tail, etc.) for each bird image. 
% Attribute annotations, however, is quite noisy. For this reason, attributes are denoised by considering class-level annotations \citep{koh2020concept}\footnote{A certain attribute is set as present only if it is also present in at least 50\% of the images of the same class. Furthermore we only considered attributes present in at least 10 classes after this refinement.}. In the end, a total of $108$ attributes (i.e. concepts with binary activations belonging to $C$) have been retained. 
The mapping $X \rightarrow C$ %from images to attribute concepts 
is performed with a ResNet10 model $g$ trained from scratch while the classifier $f$ learns the final function $C \rightarrow Y$.

\paragraph{Quantitative metrics.}
%\label{sec:exp_metrics}
% Six metrics are used to measure the performance of the proposed approach with respect to state-of-the-art methods. 
Measuring the classification quality is of crucial importance for models that are going to be applied in real-world environments. On the other hand, assessing the quality of the explanations is required for a safe%their lawful
deployment. In contrast with other kind of explanations, logic-based formulas can be evaluated quantitatively. Given a classification problem, first a set of rules are extracted for each target category from each considered model. Each explanation is then tested on an unseen set of test samples.
The results for each metric are reported in terms of mean and standard error, computed over a 5-fold cross validation \citep{krzywinski2013error}. 
% Only in the CUB experiments a 5-fold cross validation is performed due to timing issues related to BRL (each training of BRL on CUB lasts about 3 hours).
% \citep{andrews1995survey}.
For each experiment and for each model model ($f: C \rightarrow Y$ mapping concepts to target categories) six quantitative metrics are measured. 
(i) The \textsc{model accuracy} measures how well the \pb{explainer} identifies the target classes on unseen data (see Table \ref{tab:model-accuracy}).
(ii) The \textsc{explanation accuracy} measures how well the extracted logic formulas identifies the target classes (Fig.~\ref{fig:multi-objective}). This metric is obtained as the average of the F1 scores computed for each class explanation.
(iii) The \textsc{complexity of an explanation} 
%estimates how hard to understand the logic formula is for a human being (see Table \ref{tab:complexity}). This metric 
is computed by standardizing the explanations in DNF and then by counting the number of terms of the standardized formula (Fig.~\ref{fig:multi-objective}): the longer the formula, the harder the interpretation for a human being.
(iv) The \textsc{fidelity of an explanation} measures how well %class membership predictions obtained using 
the extracted explanation matches the predictions obtained using the \pb{explainer} (Table \ref{tab:fidelity}). 
% reports out-of-distribution fidelity, i.e. computed on unseen test data.
(v) The \textsc{rule extraction time} measures the time required to obtain an explanation from scratch (see Fig.~\ref{fig:time}),  computed as the sum of the time required to train the model and to extract the formula from a trained \pb{explainer}. 
%This is justified by the fact that for some models, like BRL and decision trees, training and rule extraction consist of just one simultaneous process.
(vi) The \textsc{consistency of an explanation} measures the average similarity of the extracted explanations over the \pb{5-fold cross validation runs} (see Table \ref{tab:consistency}), computed by counting how many times the same concepts appear in a logic formula over different iterations.
%folds of a cross-validation or over different initialization seeds.
% \end{itemize}

\subsection{Results and discussion}
\label{sec:results}

\begin{figure}[t]
    \centering
    \includegraphics[width=0.528\columnwidth,trim=20 20 20 20,clip]{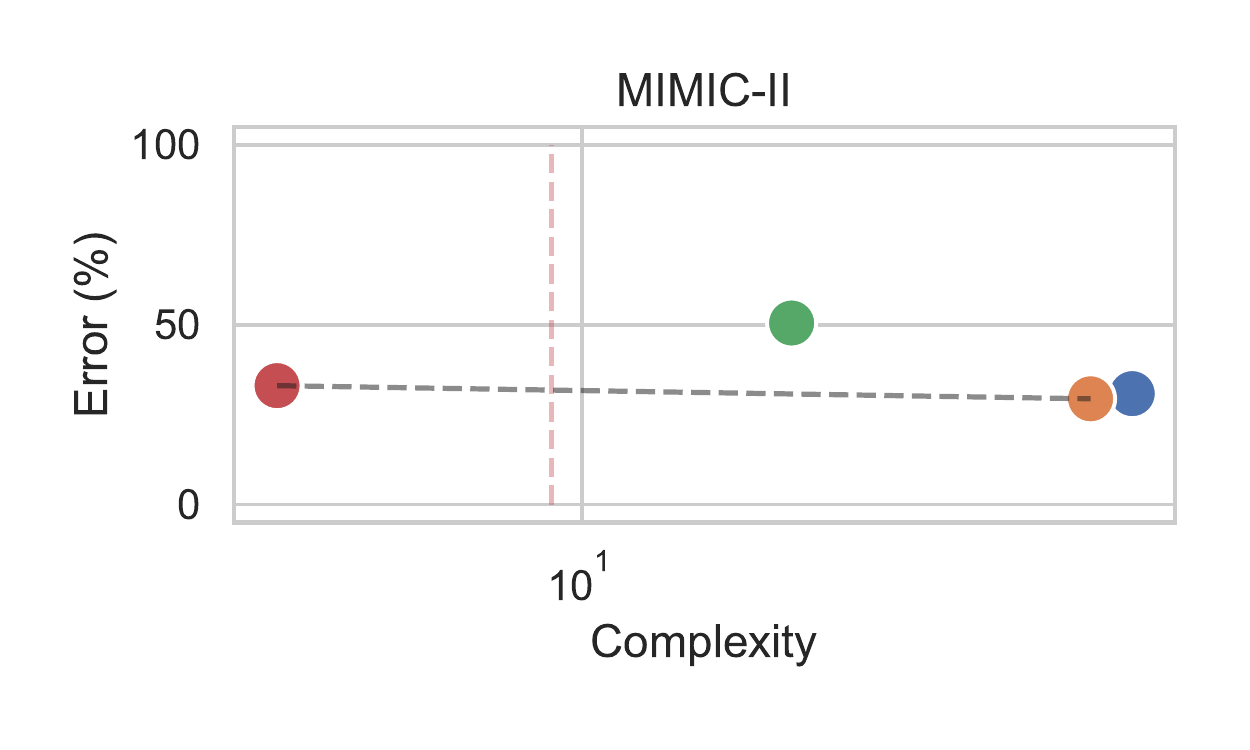}
    \includegraphics[width=0.462\columnwidth,trim=60 20 20 20,clip]{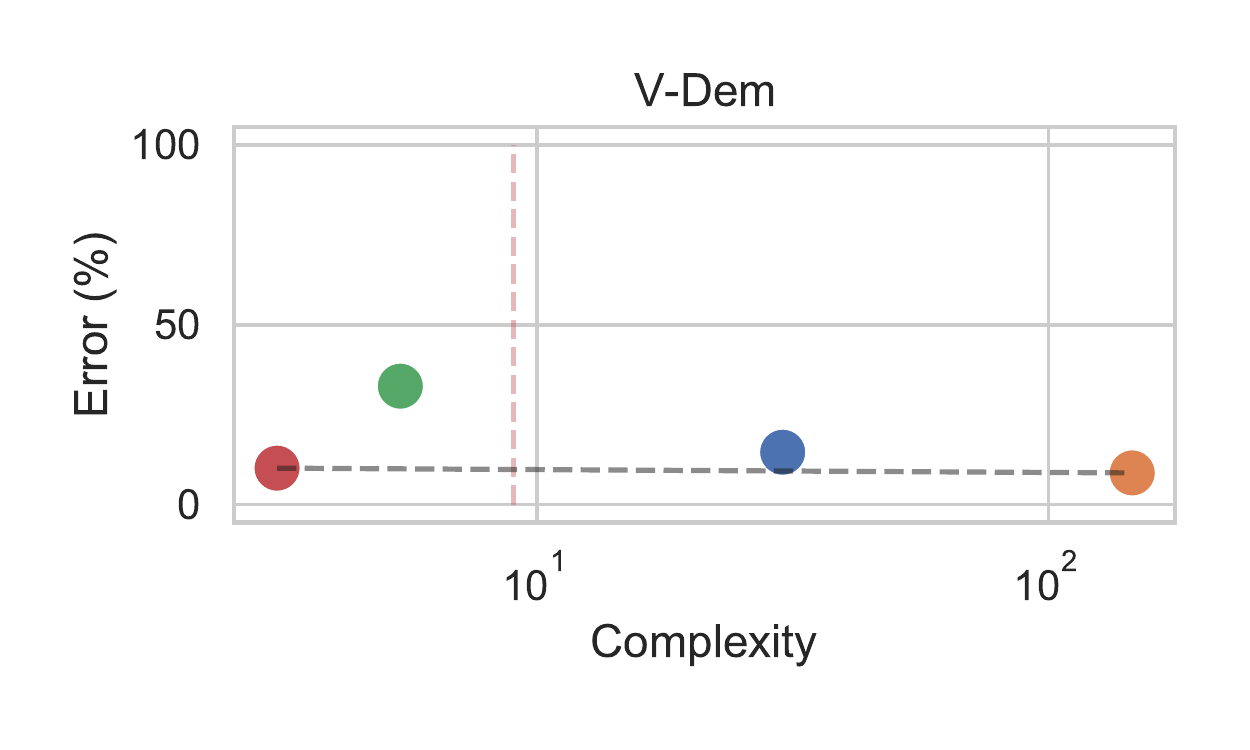}\\
    \includegraphics[width=0.528\columnwidth,trim=20 20 20 20,clip]{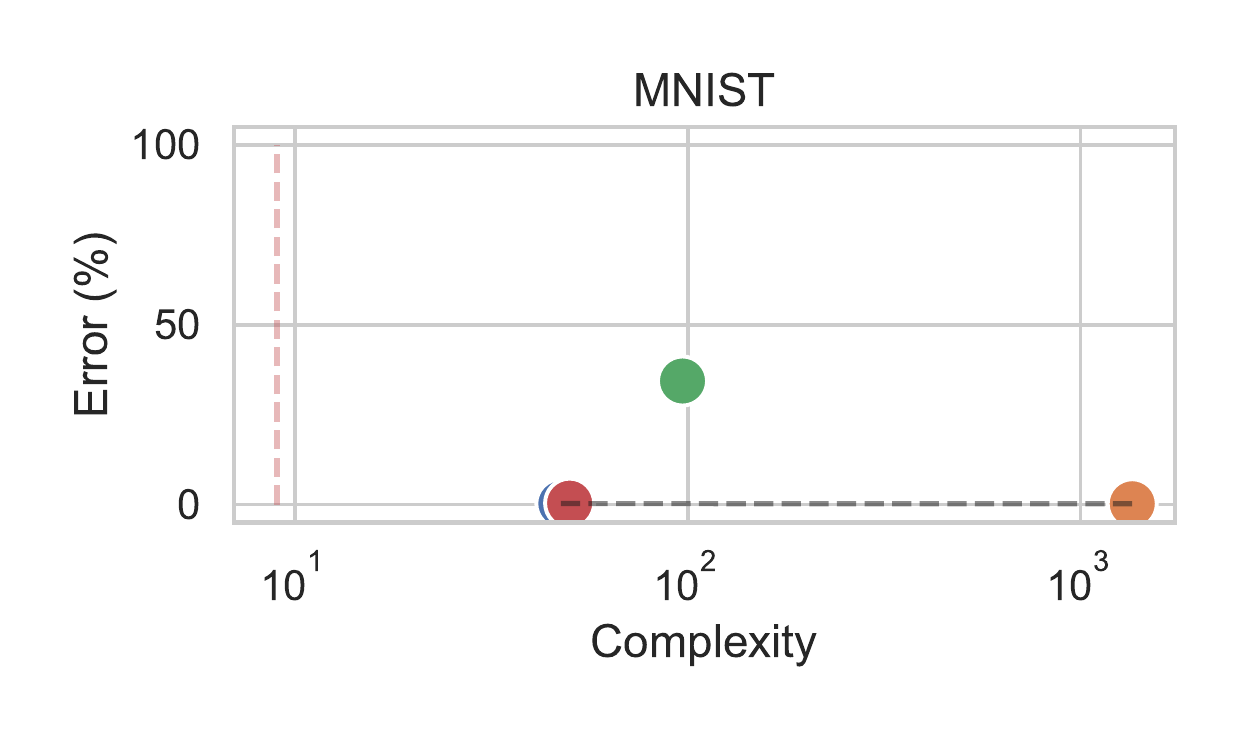}
    \includegraphics[width=0.462\columnwidth,trim=60 20 20 20,clip]{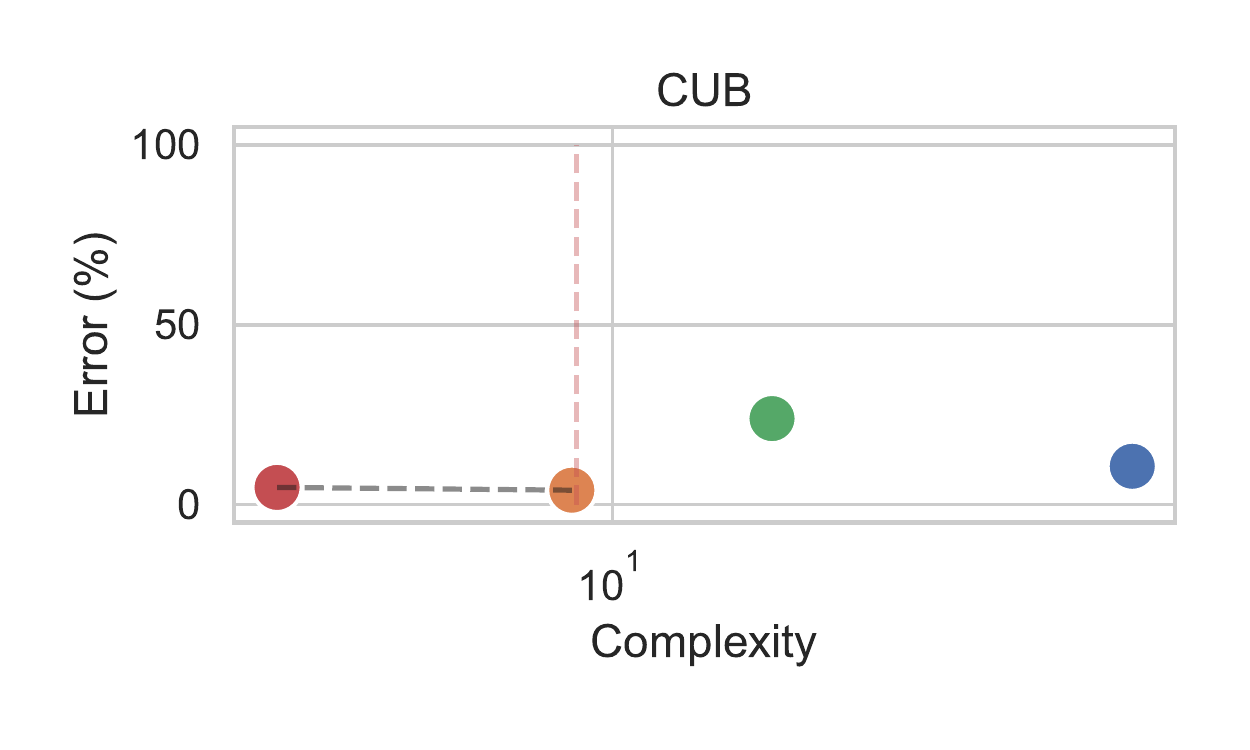}
    % \vskip -2mm    
    \includegraphics[width=\columnwidth]{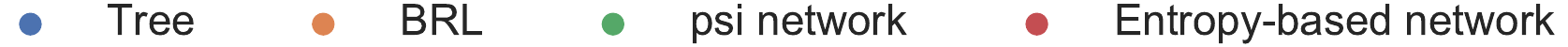}\\
    \caption{Non-dominated solutions  \citep{marler2004survey} (dotted black line) in terms of average explanation complexity and average explanation test error. The vertical dotted red line marks the maximum explanation complexity laypeople can handle (i.e. complexity $\approx 9$, see  \citep{miller1956magical,cowan2001magical,ma2014changing}). Notice how the explanations provided by the Entropy-based Network are always one of the non-dominated solution.
    % When humans compare a set of hypotheses outlining the same outcomes, they tend to have an implicit bias towards the simplest ones, making explanations from entropy-based networks the best choice.
    }
    \vskip -1mm
    \label{fig:multi-objective}
\end{figure}

\begin{figure}[t]
    \centering
    \includegraphics[width=\columnwidth,trim=30 15 35 20,clip]{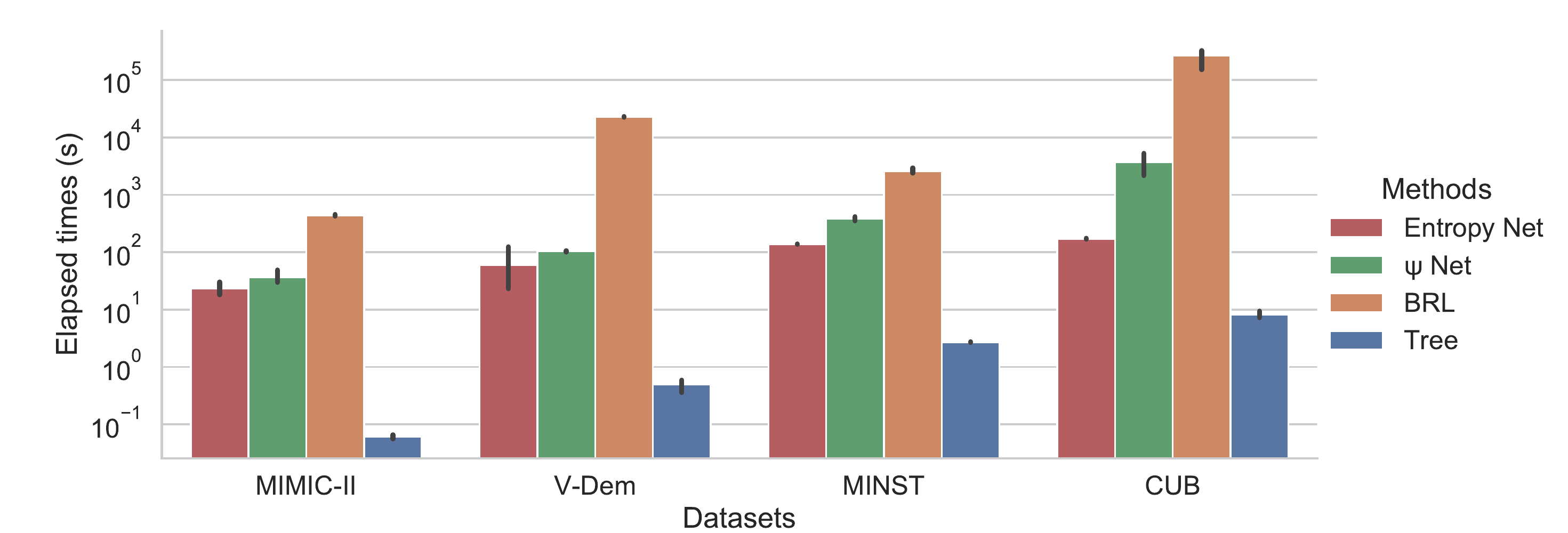}
    \caption{Time required to train models and to extract the explanations. Our model compares favorably with the competitors, with the exception of Decision Trees. BRL is by one to three order of magnitude slower than our approach.}
    % Error bars show the 95\% confidence interval of the mean.
    \label{fig:time}
    \vskip -1mm
\end{figure}

Experiments show how entropy-based networks outperform state-of-the-art white box models such as BRL and decision trees\footnote{The height of the tree is limited to obtain rules of comparable lengths. See supplementary materials \ref{appendix:exp_details}.} and interpretable neural models such as $\psi$ networks on challenging classification tasks (Table \ref{tab:model-accuracy}). 
\pb{Moreover, the entropy-based regularization and the adoption of a concept-based neural network have minor affects on the classification accuracy of the explainer when compared to
% the proposed architecture matches (and sometimes surpasses) the classification accuracy provided 
a standard black box neural network\footnote{%\pb{This is a neural network having the same architecture and hyperparameters of the entropy-based network, with the only exception of the weight hyperparameter $\lambda$ in the loss function (see Eq. \ref{eq:loss}) which is set to $\lambda=0$. This setting makes the network free from any constraint related to explainability.}
In the case of MIMIC-II and V-Dem, this is a standard neural network with the same hyperparameters of the entropy-based one, but with a linear layer as first layer. In the case of MNIST and CUB, it is the $g$ model directly predicting the final classes $g:X\rightarrow Y$.} directly working on the input data, and a Random Forest model applied on the concepts.}% as shown in Table \ref{tab:model-accuracy}.}
At the same time, the logic explanations provided by entropy-based networks are better than $\psi$ networks and almost as accurate as the rules found by decision trees and BRL, while being far more concise, as demonstrated in Fig.~\ref{fig:multi-objective}. 
More precisely, logic explanations generated by the proposed approach represent non-dominated solutions \citep{marler2004survey} \textit{quantitatively} measured in terms of complexity and classification error of the explanation. %(i.e. $100$ minus the classification accuracy of the explanation).
Furthermore, the time required to train entropy-based networks is only slightly higher with respect to Decision Trees but is lower than $\psi$ Networks and BRL by one to three orders of magnitude (Fig. \ref{fig:time}), making it feasible for explaining also complex tasks. 
% In addition, we observe how the proposed approach consistently outperform $\psi$ networks across all the main metrics (i.e. classification accuracy, explanation accuracy, and fidelity). 
The fidelity (Table~\ref{tab:fidelity})\footnote{We did not compute the fidelity of decision trees and BRL as they are trivially rule-based models.} of the formulas extracted by the entropy-based network is always higher than $90\%$ with the only exception of MIMIC. This means that almost any prediction made using the logic explanation matches the corresponding prediction made by the model, making the proposed approach very close to a white box model.
The combination of these results empirically shows that our method represents a viable solution for a safe %the lawful 
deployment of \textit{explainable} cutting-edge models.
% The complexity of decision tree formulas is never below 100 terms, making them useless as explanations.
% In terms of fidelity the proposed approach closed the gap from white-box models like decision trees and BRL, whose fidelity is always 100\% \textit{by design}.

\begin{table}[t]
\small
\centering
\caption{Out-of-distribution fidelity (\%)}
\vspace{-3mm}
\label{tab:fidelity}
%\parbox{.49\linewidth}{
\begin{tabular}{lll}
\toprule
{} &         Entropy net  &        $\psi$ net \\
\midrule
\textbf{MIMIC-II     } &  ${\bf 79.11  \pm 2.02}$ &   $51.63 \pm 6.67$ \\
\textbf{V-Dem         }&  ${\bf 90.90 \pm 1.23}$ &  $69.67 \pm 10.43$ \\
\textbf{MNIST}         &  ${\bf 99.63 \pm 0.00}$ & $65.68 \pm 5.05$ \\
\textbf{CUB         }  &  ${\bf 99.86 \pm 0.01}$ &  $77.34 \pm 0.52$ \\
\bottomrule
\end{tabular}
\vspace{-1mm}
\end{table}

\begin{table}[t]
\small
\centering
\caption{Consistency (\%)}
\vspace{-3mm}
\label{tab:consistency}
\begin{tabular}{lllll}
\toprule
{} & Entropy net     &              Tree &               BRL &        $\psi$ net\\
\midrule
\textbf{MIMIC-II     } &  $28.75$ &    ${\bf 40.49}$ &   $30.48$ &    $     27.62$ \\
\textbf{V-Dem         }&  $46.25$ &    $72.00$ &   ${\bf 73.33}$ &    $     38.00$ \\
\textbf{MNIST}         &  ${\bf 100.00}$ &   $41.67$ &  ${\bf 100.00}$ &    $96.00$ \\
\textbf{CUB         }  &  $35.52$ &    $21.47$ &   ${\bf 42.86}$ &    $41.43$ \\
\bottomrule
\end{tabular}
\vspace{-2mm}
\end{table}

The reason why the proposed approach consistently outperform $\psi$ networks across all the key metrics (i.e. classification accuracy, explanation accuracy, and fidelity) can be explained observing how entropy-based networks are far less constrained than $\psi$ networks, both in the architecture (our approach does not apply weight pruning) and in the loss function (our approach applies a regularization on the distributions $\alpha^i$ and not on all weight matrices). Likewise, the main reason why the proposed approach provides a higher classification accuracy with respect to BRL and decision trees may lie in the smoothness of the decision functions of neural networks which tend to generalize better than rule-based methods, as already observed by Tavares et al. \citep{tavares2020understanding}.
For each dataset, we report in the supplemental material (Appendix \ref{appendix:formulas}) a few examples of logic explanations extracted by each method, as well as in Fig. \ref{fig:experiments}. We mention that the proposed approach is the only matching the logically correct ground-truth explanation for the MNIST even/odd experiment, i.e. $\forall x, \mathrm{isOdd(x)} \leftrightarrow \mathrm{isOne(x)} \oplus \mathrm{isThree(x)} \oplus \mathrm{isFive(x)} \oplus \mathrm{isSeven(x)} \oplus \mathrm{isNine(x)}$ and $\forall x, \mathrm{isEven(x)} \leftrightarrow \mathrm{isZero(x)} \oplus \mathrm{isTwo(x)} \oplus \mathrm{isfour(x)} \oplus \mathrm{isSix(x)} \oplus \mathrm{isEight(x)}$, being $\oplus$ the exclusive OR.
In terms of formula consistency, we observe how BRL is the most consistent rule extractor, closely followed by the proposed approach (Table \ref{tab:consistency}).

\section{Conclusions} \label{sec:conclusion}
This work contributes to a %lawful and 
safer adoption of some of the most powerful AI technologies, allowing deep neural networks to have a greater impact on society by making them explainable-by-design, thanks to an entropy-based approach that yields FOL-based explanations.
Moreover, as the proposed approach provides logic explanations for how a model arrives at a decision, it can be effectively used to reverse engineer algorithms, processes, to find vulnerabilities, or to improve system design powered by \pb{deep learning models}. 
From a scientific perspective, formal knowledge distillation from state-of-the-art networks may enable scientific discoveries or falsification of existing theories.
However, the extraction of a FOL explanation requires symbolic input and output spaces. In some contexts, such as computer vision, the use of concept-based approaches may require additional annotations and attribute labels to get a consistent symbolic layer of concepts. Recent works on automatic concept extraction may alleviate the related costs, leading to more cost-effective concept annotations \citep{ghorbani2019towards,kazhdan2020now}.

\section*{Acknowledgments and Disclosure of Funding}
We thank Ben Day, Dobrik Georgiev, Dmitry Kazhdan, and Alberto Tonda for useful feedback and suggestions.

This work was partially supported by TAILOR and GODS-21 European Union’s Horizon 2020 research and innovation programmes under GA No 952215 and 848077.

% This work was partially supported by the European Union’s Horizon 2020 research and innovation programme under grant agreement No .
%  by TAILOR, a project funded by EU Horizon 2020 research and innovation programme under GA No .”
% \fi

{\small
\bibliography{aaai22}
}

\clearpage

\appendix
\section{Appendix}
\label{appendix}

\subsection{Software}
\label{appendix:software}
In order to make the proposed approach accessible to the whole community, we released 
Anonymous \citep{anonymous}, % "PyTorch, Explain!" \citep{barbiero2021lens}
a Python package\footnote{\url{https://github.com/pietrobarbiero/entropy-lens}}
with an extensive documentation on methods and unit tests. The Python code and the scripts used for the experiments, including parameter values and documentation, is freely available under Apache 2.0 Public License from a GitHub repository.

The code library is designed with intuitive APIs requiring only a few lines of code to train and get explanations from the neural network as shown in the following code snippet \ref{code:example}.

\begin{lstlisting}[language=Python, label=code:example, caption=Example on how to use the APIs to implement the proposed approach.]
import torch_explain as te
from torch_explain.logic import test_explanation
from torch_explain.logic.nn import explain_class

# XOR problem with additional features
x0 = torch.zeros((4, 100))
x = torch.tensor([
    [0, 0, 0],
    [0, 1, 0],
    [1, 0, 0],
    [1, 1, 0],
], dtype=torch.float)
x = torch.cat([x, x0], dim=1)
y = torch.tensor([0, 1, 1, 0], dtype=torch.long)

# network architecture
layers = [
    te.nn.EntropyLogicLayer(x.shape[1], 10, n_classes=2),
    torch.nn.LeakyReLU(),
    te.nn.LinearIndependent(10, 10, n_classes=2),
    torch.nn.LeakyReLU(),
    te.nn.LinearIndependent(10, 1, n_classes=2, top=True)
]
model = torch.nn.Sequential(*layers)

# train loop
optimizer = torch.optim.AdamW(model.parameters(), lr=0.01)
loss_form = torch.nn.CrossEntropyLoss()
model.train()
for epoch in range(1001):
    optimizer.zero_grad()
    y_pred = model(x)
    loss = loss_form(y_pred, y) + \
           0.00001 * te.nn.functional.entropy_logic_loss(model)
    loss.backward()
    optimizer.step()

# logic explanations
y1h = one_hot(y)
_, class_explanations, _ = explain_class(model, x, y1h, x, y1h)

\end{lstlisting}

% # instantiate a mu network
% model = dl.models.MuNN(n_classes=n_classes, n_features=n_features,
%                       hidden_neurons=[200], loss=torch.nn.CrossEntropyLoss(), 
%                       l1_weight=0.001, fan_in=10)

% # fit the model
% model.fit(train_data, val_data, epochs=100, l_r=0.0001)

% # get predictions on test samples
% outputs, labels = model.predict(test_data)

% # get first-order logic explanations for a specific target class
% target_class = 1
% formula = model.get_global_explanation(x_val, y_val, target_class)
% # compute explanation accuracy
% accuracy = dl.logic.test_explanation(formula, target_class, x_test, y_test)

\subsection{Dataset Description}
\label{appendix:dataset}
%\vspace{-3mm}
\paragraph{Will we recover from ICU? (MIMIC-II).}
The Multiparameter Intelligent Monitoring in Intensive Care II (MIMIC-II, \citep{saeed2011multiparameter,goldberger2000physiobank}) is a public-access intensive care unit (ICU) database consisting of 32,536 subjects (with 40,426 ICU admissions) admitted to different ICUs.
%medical, surgical, cardiovascular and neonatal ICUs, surgical recovery units, and coronary care units ICUs at a single tertiary care hospital. 
The dataset contains detailed descriptions of a variety of clinical data classes: general,
%(e.g. patient demographics, hospital admission and discharge dates, hospital death dates, etc), 
physiological, % (e.g.	nurse-verified vital signs, blood pressure, heart rate, etc), 
results of clinical laboratory tests, % (e.g. hematology, blood chemistries, urinalysis, microbiology, etc), 
records of medications, fluid balance, %(e.g. solutions, urine, estimated blood loss, etc), 
and text reports of imaging studies (e.g. x-ray, CT, MRI, etc). In our experiments, we removed non-anonymous information, text-based features, time series inputs, and observations with missing data.
%discretize some of the continuous features into a set of categories in order to get a one-hot encoded representation. For instance, 
We discretize continuous features
% , like blood pressure (BP), 
into one-hot encoded categories.
% (e.g. very low BP, low BP, normal BP, high BP, and very high BP). 
After such preprocessing step, we obtained an input space $C$ composed of $k=90$ key features. 
The task consists in 
% training a classifier function to 
identifying recovering %($Y=1$) 
or dying %($Y=0$) 
patients after ICU admission. 
%An end-to-end classifier $f: C \rightarrow Y$ is employed to carry out the classification task.

%\vspace{-3mm}
\paragraph{What kind of democracy are we living in? (V-Dem).}
Varieties of Democracy  (V-Dem, \citep{pemstein2018v,coppedge2021v}) is a dataset containing a collection of indicators of latent regime characteristics over 202 countries from 1789 to 2020. The database include $k_1 = 483$ low-level indicators (e.g. media bias, party ban, high-court independence,  etc.), $k_2 = 82$ mid-level indices (e.g. freedom of expression, freedom of association, equality before the law, etc), and 5 high-level indices of democracy principles (i.e. electoral, liberal, participatory, deliberative, and egalitarian). In the experiments a binary classification problem is considered to identify electoral democracies %($Y=1$) 
from non-electoral democracies. %($Y=0$). 
We indicate with $C_1$ and $C_2$ the spaces associated to the activations of the aforementioned two levels of concepts. Two classifiers $f_1$ and $f_2$ are trained to learn the map $C_1 \rightarrow C_2 \rightarrow Y$. Explanations are given for classifier $f_2$ in terms of concepts $c_2 \in C_2$.

% \paragraph{MNIST Even/Odd}
%\vspace{-3mm}
\paragraph{What does parity mean? (MNIST Even/Odd).}
The Modified National Institute of Standards and Technology database (MNIST, \citep{lecun1998mnist}) contains a large collection of images representing handwritten digits. The input space $X \subset \mathbb{R}^{28\times 28}$ is composed of 28x28 pixel images while the concept space $C$ with $k=10$ is represented by the label indicator for digits from $0$ to $9$. 
% However, 
The task we consider here is slightly different from the common digit-classification. Assuming $Y \subset \{ 0,1 \}^2$, we are interested in determining if a digit is either odd or even, and explaining the assignment to one of these classes in terms of the digit labels (concepts in $C$). 
% Notice how, for this classification problem, we trivially have ground-truth first-order logic formulas: $\forall x, \mathrm{isOdd(x)} \leftrightarrow \mathrm{isOne(x)} \oplus \mathrm{isThree(x)} \oplus \mathrm{isFive(x)} \oplus \mathrm{isSeven(x)} \oplus \mathrm{isNine(x)}$ and $\forall x, \mathrm{isEven(x)} \leftrightarrow \mathrm{isZero(x)} \oplus \mathrm{isTwo(x)} \oplus \mathrm{isfour(x)} \oplus \mathrm{isSix(x)} \oplus \mathrm{isEight(x)}$, being $\oplus$ the exclusive OR. 
The mapping $X \rightarrow C$ is provided by a ResNet10 classifier $g$ \citep{he2016deep} trained from scratch.
while the classifier $f$ is used to learn both the final mapping and the explanation as a function $C \rightarrow Y$.

%\vspace{-3mm}
\paragraph{What kind of bird is that? (CUB).}
The Caltech-UCSD Birds-200-2011 dataset (CUB, \citep{wah2011caltech}) is a fine-grained classification dataset. It includes 11,788 images representing $r = 200$ ($Y = \{0,1\}^{200}$) different bird species. 312 binary attributes describe visual characteristics (color, pattern, shape) of particular parts (beak, wings, tail, etc.) for each bird image. 
Attribute annotations, however, is quite noisy. For this reason, attributes are denoised by considering class-level annotations \citep{koh2020concept}\footnote{A certain attribute is set as present only if it is also present in at least 50\% of the images of the same class. Furthermore we only considered attributes present in at least 10 classes after this refinement.}. In the end, a total of $108$ attributes (i.e. concepts with binary activations belonging to $C$) have been retained. The mapping $X \rightarrow C$ from images to attribute concepts is performed again with a ResNet10 model $g$ trained from scratch while the classifier $f$ learns the final function $C \rightarrow Y$.

All datasets employed are freely available (only MIMIC-II requires an online registration) and can be downloaded from the following links: \\
MIMIC: \url{https://archive.physionet.org/mimic2}. \\
V-Dem: \url{https://www.v-dem.net/en/data/data/v-dem-dataset-v111}. \\
MNIST: \url{http://yann.lecun.com/exdb/mnist}. \\
CUB: \url{http://www.vision.caltech.edu/visipedia/CUB-200-2011.html}.

\subsection{Experimental details}
\label{appendix:exp_details}

Batch gradient-descent and the Adam optimizer with decoupled weight decay \citep{loshchilov2017decoupled} and learning rate set to $10^{-2}$ are used for the optimization of all neural models' parameters (Entropy-based Network and $\psi$ Network). An early stopping strategy is also applied: the model with the highest accuracy on the validation set is saved and restored before evaluating the test set.

With regard to the Entropy-based Network, Tab. \ref{tab:hyperparameters} reports the hyperparameters employed  to train the network in all experiments. All Entropy-based Networks feature ReLU activations and linear fully-connected layers (except for the first layer which is the Entropy Layer). A grid search cross-validation strategy has been employed on the validation set to select hyperparameter values. The objective was to maximize at the same time both model and explanation accuracy. $\lambda$ represents the trade-off parameter in Eq. \ref{eq:loss} while $\tau$ is the temperature of Eq. \ref{eq:alpha}.

\begin{table}[!ht]
\centering
\caption{Hyper parameters of entropy-based networks.}
\label{tab:hyperparameters}
% \resizebox{\textwidth}{!}{
\begin{tabular}{lllll}
\toprule
{} &         $\lambda$    & $\tau$ &  max epochs & hidden neurons \\
\midrule
\textbf{MIMIC-II} & $10^{-3}$ &  $0.7$  &  200 & $20$ \\
\textbf{V-Dem} & $10^{-5}$ &  $5$       &  200 & $20, 20$ \\
\textbf{MNIST} & $10^{-7}$ &  $5$       &  200 & $10$ \\
\textbf{CUB} & $10^{-4}$ &  $0.7$       &  500 & $10$ \\

\bottomrule
\end{tabular}
% }
\end{table}

Concerning the $\psi$ network in all experiments one network per class has been trained. They are composed of two hidden layer of 10 and 5 hidden neurons respectively. As indicated in the original paper, an $l_1$ weight regularization has been applied to all layers of the network. As in this work, the contribute in the overall loss of the $l_1$ regularization is weighted by an hyperparameter $\lambda = 10^{-4}$. The maximum number of non-zero input weight (fan-in) is set to 3 in in MIMIC and V-Dem while for MNIST and CUB200 it is set to 4. In Ciravegna et al. \citep{ciravegna2020human}, $\psi$ networks were devised to provide explanations of existing models; in this paper, however, we have shown how they can directly solve classification problems.

Decision Trees have been limited in their maximum height in all experiments to maintain the complexity of the rules at a comparable level w.r.t the other methods. More precisely the maximum height has been set to $5$ in all binary classification tasks (MIMIC-II, V-Dem, MNIST) while we allowed a maximum height of $30$ in the CUB experiment due to the high number of classes to predict (200).

BRL algorithms requires to first run the FP-growth algorithm \citep{han2000mining} (an enhanced version of Apriori) to mine a first set of frequent rules. The hyperparameter used by FP-growth are: the minimum support in percentage of training samples for each rule (set to $10\%$), the minimum and the maximum number of features considered by each rule (respectively set to 1 and 2). Regarding the Bayesian selection of the best rules, the number of Markov chain Monte Carlo used for inference is set to 3, while 50000 iterations maximum are allowed. At last the expected length and width of the extracted rule list is set respectively to 3 and 1. These are the default values indicated in the BRL repository. Due to the computational complexity and the high number of hyperparameters, they have not been cross validated. 
 
The code for the experiments is implemented in Python 3, relying upon open-source libraries \citep{paszke2019pytorch,pedregosa2011scikit}.
All the experiments have been run on the same machine: Intel\textsuperscript{\textregistered} Core\texttrademark\ i7-10750H 6-Core Processor at 2.60 GHz equipped with 16 GiB RAM and NVIDIA GeForce RTX 2060 GPU.

% \clearpage
\subsection{Explainability metrics details}
\label{appendix:results}
In the following, we report  
% the out-of-distribution fidelity (Table \ref{tab:fidelity}) and rule consistency (Table \ref{tab:consistency}) for all the benchmark datasets presented in the experimental section. 
% We also report 
in tabular form the results concerning the explanation accuracy and the complexity of the rules (Fig. \ref{fig:multi-objective}) and the extraction time (Fig. \ref{fig:time}).

% \begin{table}[!ht]
% \begin{minipage}{0.44\textwidth}
% \begin{table}[!ht]
% \centering
% \caption{Out-of-distribution fidelity (\%)}
% \label{tab:fidelity}
% \footnotesize
% %\parbox{.49\linewidth}{
% \begin{tabular}{lll}
% \toprule
% {} &         Entropy net  &        $\psi$ net \\
% \midrule
% \textbf{MIMIC-II     } &  ${\bf 79.11  \pm 2.02}$ &   $51.63 \pm 6.67$ \\
% \textbf{V-Dem         }&  ${\bf 90.90 \pm 1.23}$ &  $69.67 \pm 10.43$ \\
% \textbf{MNIST}         &  ${\bf 99.63 \pm 0.00}$ & $65.68 \pm 5.05$ \\
% \textbf{CUB         }  &  ${\bf 99.86 \pm 0.01}$ &  $77.34 \pm 0.52$ \\
% \bottomrule
% \end{tabular}
% \end{table}

% \begin{table}[!ht]
% \small
% \centering
% \caption{Consistency (\%)}
% \label{tab:consistency}
% \footnotesize
% \begin{tabular}{lllll}
% \toprule
% {} & Entropy net     &              Tree &               BRL &        $\psi$ net\\
% \midrule
% \textbf{MIMIC-II     } &  $28.75$ &    ${\bf 40.49}$ &   $30.48$ &    $     27.62$ \\
% \textbf{V-Dem         }&  $46.25$ &    $72.00$ &   ${\bf 73.33}$ &    $     38.00$ \\
% \textbf{MNIST}         &  ${\bf 100.00}$ &   $41.67$ &  ${\bf 100.00}$ &    $96.00$ \\
% \textbf{CUB         }  &  $35.52$ &    $21.47$ &   ${\bf 42.86}$ &    $41.43$ \\
% \bottomrule
% \end{tabular}
% \end{table}

\begin{table}[!ht]
\centering
\caption{Explanation's accuracy (\%) computed as the average of the F1 scores computed for each class.}
\label{tab:explanation-accuracy}
\resizebox{\columnwidth}{!}{
\begin{tabular}{lllll}
\toprule
{} &         Entropy net    &        Tree &              BRL &               $\psi$ net  \\
\midrule
\textbf{MIMIC-II     } &  $66.93 \pm 2.14$  &  $69.15 \pm 2.24$ &  ${\bf 70.59 \pm 2.17}$ &  $49.51 \pm 3.91$ \\
\textbf{V-Dem         } &  $89.88 \pm 0.50$ &  $85.45 \pm 0.58$ &  ${\bf 91.21 \pm 0.75}$ &  $67.08 \pm 9.68$ \\
\textbf{MNIST}         &  $99.62 \pm 0.00$ &  $99.74 \pm 0.01$ &  ${\bf 99.79 \pm 0.02}$ &  $65.64 \pm 5.05$ \\
\textbf{CUB         } &  $95.24 \pm 0.05$  &   $89.36 \pm 0.92$ &  ${\bf 96.02 \pm 0.17}$ &  $76.10 \pm 0.56$ \\

\bottomrule
\end{tabular}
}
\end{table}

\begin{table}[!ht]
\centering
\caption{Complexity computed as the number of terms in each minterm of the DNF rules.}
\label{tab:complexity}
\resizebox{\columnwidth}{!}{
\begin{tabular}{lllll}
\toprule
{} &         Entropy net    &      Tree &              BRL  &               $\psi$ net  \\
\midrule
\textbf{MIMIC-II     } &  ${\bf 3.50 \pm 0.88}$ &   $66.60 \pm 1.45$ &     $57.70 \pm 35.58$ &   $20.6 \pm 5.36$ \\
\textbf{V-Dem         } &  ${\bf 3.10 \pm 0.51}$ &  $30.20 \pm 1.20$ &    $145.70 \pm 57.93$ &   $5.40 \pm 2.70$ \\
\textbf{MNIST}         &  $50.00 \pm 0.00$ &    ${\bf 47.50 \pm 0.72}$ &  $1352.30 \pm 292.62$ &   $96.90 \pm 10.01$ \\
\textbf{CUB          } &  ${\bf 3.74 \pm 0.03}$ &    $45.92 \pm 1.16$ &       $8.87 \pm 0.11$ &   $15.96 \pm 0.96$ \\

\bottomrule
\end{tabular}
}
\end{table}

\begin{table}[!ht]
\centering
\caption{Rule extraction time (s) calculated as the time required to train the models and to extract the corresponding rules.}
\label{tab:extraction}
\resizebox{\columnwidth}{!}{
\begin{tabular}{lllll}
\toprule
{} &         Entropy net    &          Tree &               BRL &  $\psi$ net \\
\midrule
\textbf{MIMIC-II     } &  $23.08 \pm 3.53$   &  ${\bf 0.06 \pm 0.00}$ &         $440.24 \pm 9.75$ &       $36.68 \pm 6.10$ \\
\textbf{V-Dem         }&  $59.90 \pm 31.18$   &  ${\bf 0.49 \pm 0.07}$ &     $22843.21 \pm 194.49$ &       $103.78 \pm 1.65$ \\
\textbf{MNIST}         &  $138.32 \pm 0.63$   &  ${\bf 2.72 \pm 0.02}$ &      $2594.79 \pm 177.34$ &      $385.57 \pm 17.81$ \\
\textbf{CUB          } &  $171.87 \pm 1.95$ &  ${\bf 8.10 \pm 0.65}$ &  $264678.29 \pm 56521.40$ &  $3707.29 \pm 1006.54$ \\

\bottomrule
\end{tabular}
}
\end{table}

\subsection{Entropy and L1}
\label{appendix:entropy_vs_L1}

This section presents additional experiments on a toy dataset showing (1) the advantage of using the entropy loss function in Eq. \ref{eq:ent} w.r.t. the L1 loss (used by e.g. the $\psi$ network) and (2) the advantage in terms of explainability provided by the Entropy Layer w.r.t. a standard linear layer. Three neural models are compared:
\begin{itemize}
    \item model A: a standard multi-layer perceptron using linear (fully connected) layers, using an L1 regularization in the loss function.
    \item model B: a multi-layer perceptron using the Entropy Layer as first layer and an L1 regularization in the loss function.
    \item model C: a multi-layer perceptron using the Entropy Layer as first layer and the entropy loss regularization (Eq. \ref{eq:ent}) in the loss function.
\end{itemize}

The dataset used for this experiment is shown in Table \ref{tab:toy_dataset}. The training set is composed of four Boolean features $\{x_1, x_2, x_3, x_4\}$ and four Boolean target categories $\{y, \neg y, z, \neg z\}$. The target category $y$ is the XOR of the features $x_1$ and $x_2$, i.e. $\forall x: y=1 \leftrightarrow x_1 \oplus x_2$. The target category $z$ is the OR of the features $x_3$ and $x_4$, i.e. $\forall x: z=1 \leftrightarrow x_3 \wedge x_4$. The categories $\neg y$ and $\neg z$ are the complement of the categories $y$  and $z$, respectively.

The neural networks used for these experiment are multi-layer perceptrons with 2 hidden layers of $20$ and $10$ units with ReLu activation. Batch gradient-descent and the Adam optimizer with decoupled weight decay \citep{loshchilov2017decoupled} and learning rate set to $10^{-4}$ are used for all neural models. The number of epochs is set to $18000$ to ensure complete convergence (overfitting the training set), and the regularization coefficient is set to $\lambda = 10^{-4}$ for both L1 and entropy losses. For the neural model using the entropy loss (model C), the temperature is set to $\tau = 0.3$.

\begin{figure}[!ht]
    \centering
    \includegraphics[width=\columnwidth]{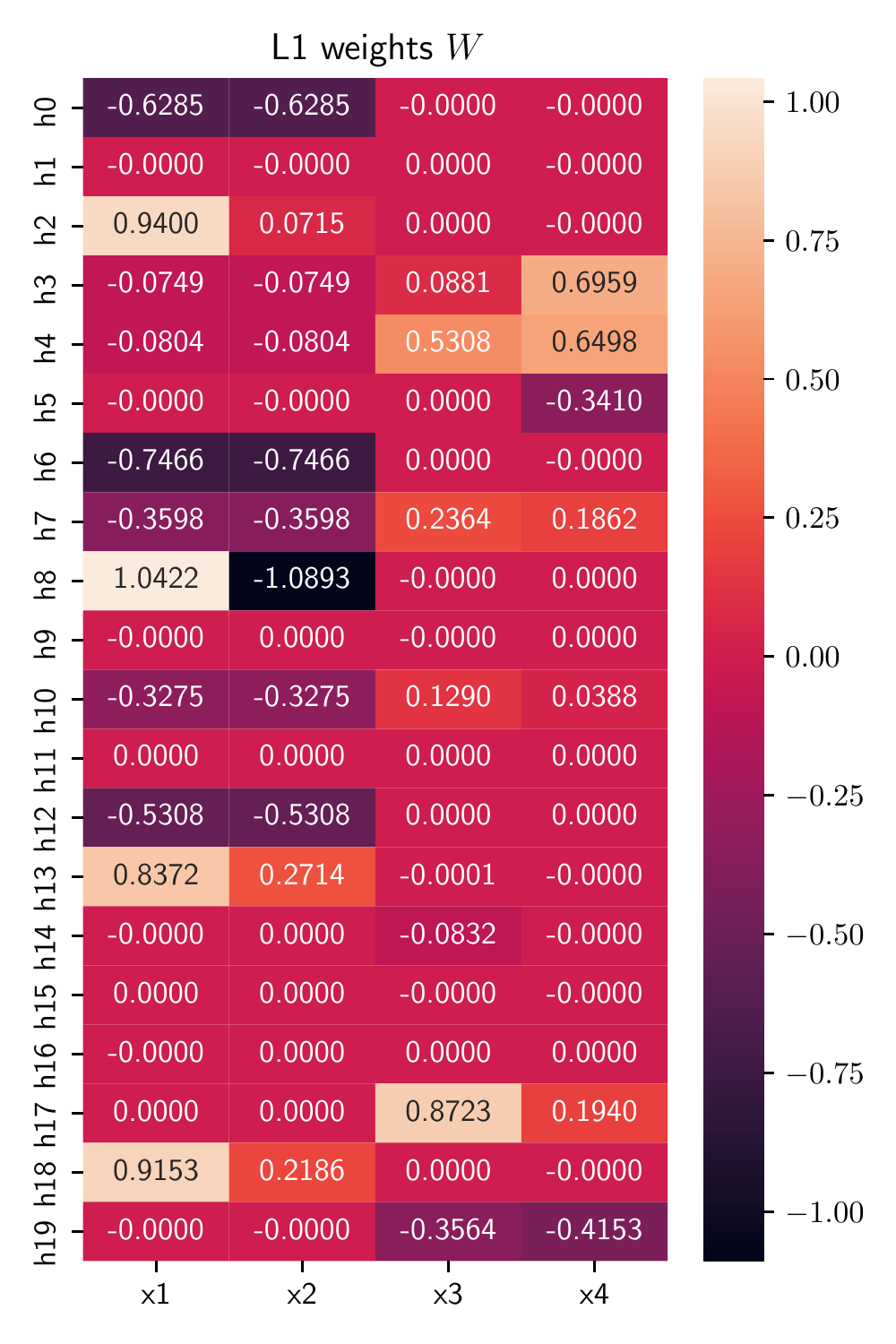}
    \caption{Concept relevance in terms of the weights $W$ of a standard linear layer (not entropy layer) used as the first layer of the network trained by minimizing the L1 loss (model A).}
    \label{fig:L1_standard}
\end{figure}

\begin{figure}[!ht]
    \centering
    \includegraphics[width=\columnwidth]{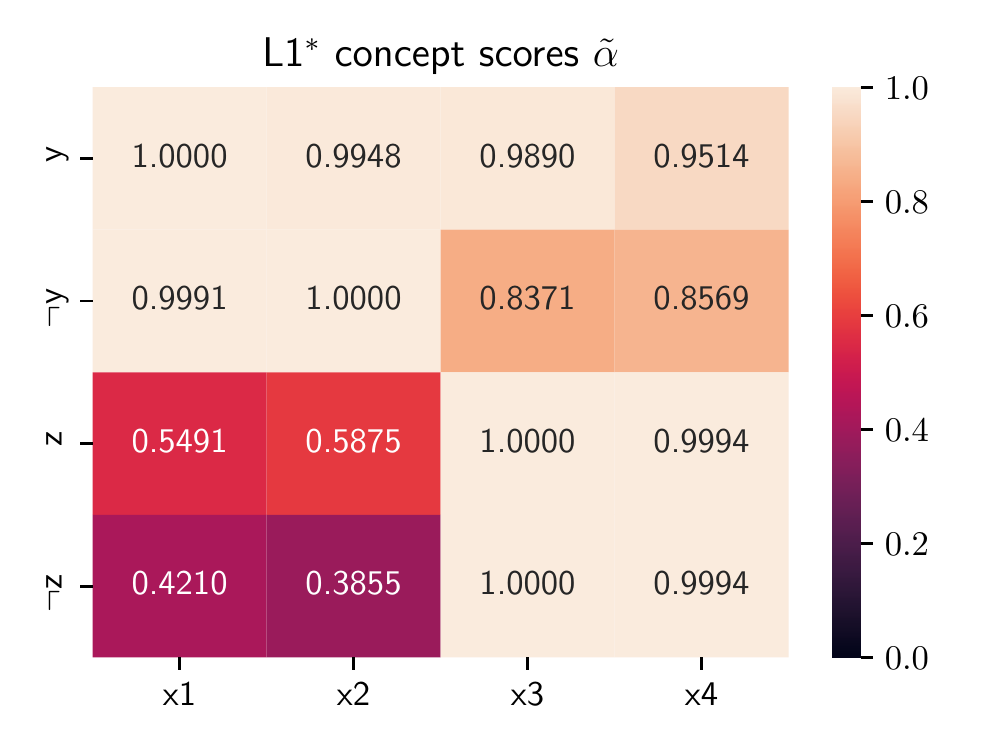}
    \caption{Concept scores $\tilde{\alpha}$ for the Entropy layer (first layer of the network) trained by minimizing the L1 loss instead of the entropy loss of Eq. \ref{eq:ent} (model B).}
    \label{fig:L1_entropy}
\end{figure}

\begin{figure}[!ht]
    \centering
    \includegraphics[width=\columnwidth]{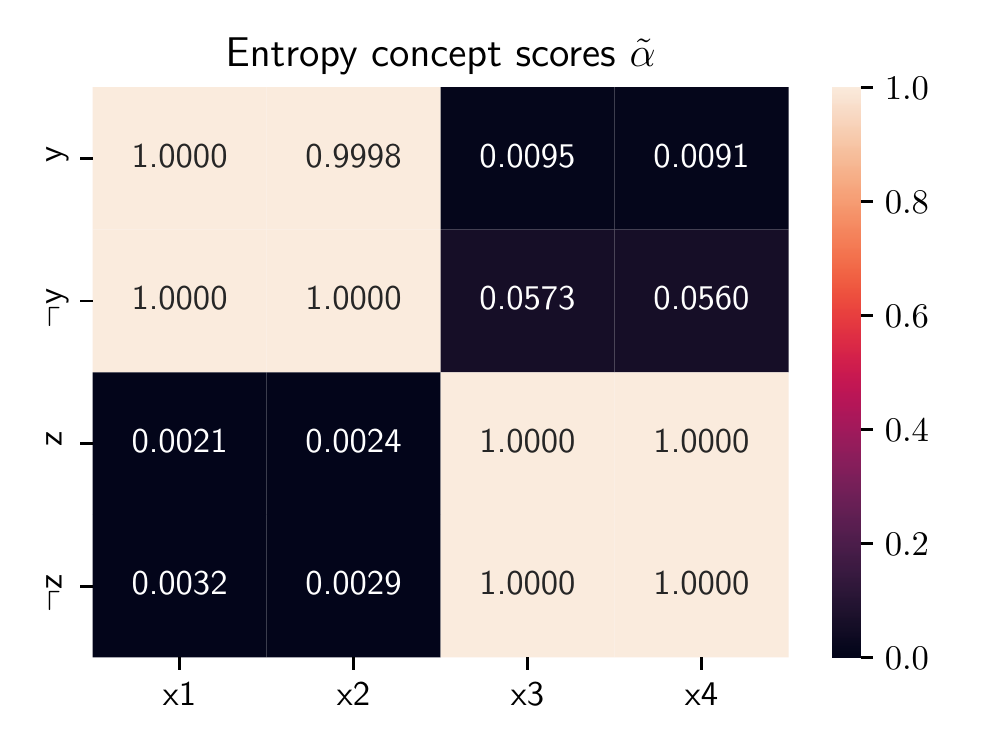}
    \caption{Concept scores $\tilde{\alpha}$ for the Entropy layer (first layer of the network) trained by minimizing the entropy loss in Eq. \ref{eq:ent} (model C).}
    \label{fig:entropy_alpha}
\end{figure}

\begin{figure*}[!ht]
    \centering
    \includegraphics[width=0.9\textwidth]{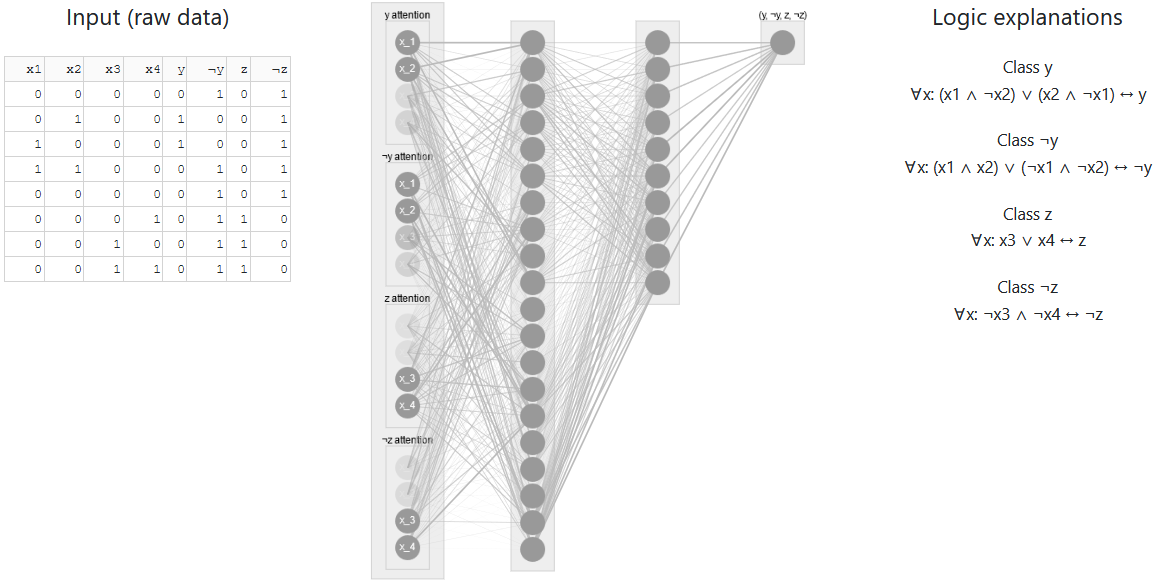}
    \caption{Visualization of the Entropy-based Network (model C) trained on the toy dataset after $18000$ epochs. The first layer of the network is the Entropy Layer. The gray-scale intensity of the input neurons represents the concept scores $\tilde{\alpha}$. The darker the input neuron, the higher the score ($\tilde{\alpha}_i^j \rightarrow 1$), the lighter the input neuron, the lower the score ($\tilde{\alpha}_i^j \rightarrow 0$).}
    \label{fig:elen_trained}
\end{figure*}

Once the networks have been trained, we extracted from each model a summary of the concept relevance for each target category. Figure \ref{fig:L1_standard} shows the values of the weight matrix of the first hidden layer of the model A (not using the Entropy Layer). The L1 loss pruned some connections between input features and hidden neurons ($h_i$). However, it is not evident the relevance of each feature for each target class. Figure \ref{fig:L1_entropy} shows the matrix $\tilde{\alpha}$ of the concept scores provided by the Entropy layer of the model B (trained with the L1 loss). It can be observed how the matrix $\tilde{\alpha}$ offers a much better overview of the relevance of each feature for each target category. However, the L1 loss was not sufficient to make the model learn that e.g. the category $y$ does not depend from the feature $x_3$ (recall that $\forall x: y=1 \leftrightarrow x_1 \oplus x_2$), as the score $\tilde{\alpha}_{x_3}^y \approx 0.99$. Finally, Figure \ref{fig:entropy_alpha} shows the matrix $\tilde{\alpha}$ of the concept scores provided by the Entropy layer of the model C (trained with the entropy loss in Eq. \ref{eq:ent}). The entropy loss was quite effective helping the neural network identify the most relevant input features for each task, discarding redundant input concepts. Figure \ref{fig:elen_trained} shows the trained Entropy Network (model C) on the toy dataset as well as the resulting logic explanations inferred from the training set matching ground-truth logic formulas.

% \usepackage{graphicx}
\begin{table}[t]
\centering
\caption{Toy dataset used to compare the Entropy Layer to a standard linear layer and the entropy loss to the L1 loss function.}
\label{tab:toy_dataset}
\resizebox{\columnwidth}{!}{%
\begin{tabular}{lllll|lllll}
$x_1$ & $x_2$ & $x_3$ & $x_4$ &  &  & $y$ & $\neg y$ & $z$ & $\neg z$ \\ \hline
0     & 0     & 0     & 0     &  &  & 0   & 1        & 0   & 1        \\
0     & 1     & 0     & 0     &  &  & 1   & 0        & 0   & 1        \\
1     & 0     & 0     & 0     &  &  & 1   & 0        & 0   & 1        \\
1     & 1     & 0     & 0     &  &  & 0   & 1        & 0   & 1        \\
0     & 0     & 0     & 0     &  &  & 0   & 1        & 0   & 1        \\
0     & 0     & 0     & 1     &  &  & 0   & 1        & 1   & 0        \\
0     & 0     & 1     & 0     &  &  & 0   & 1        & 1   & 0        \\
0     & 0     & 1     & 1     &  &  & 0   & 1        & 1   & 0       
\end{tabular}%
}
\end{table}

\subsection{Impact of hyperparameters on the logic explanations}
\label{appendix:hyperparam_explanations}

We measured the impact of the hyperparameters on the quality of the explanations on the vDem dataset. Table \ref{tab:hyperparam_explanations}  summarizes the average model accuracy, explanation accuracy, and explanation complexity obtained by running a 5-fold cross validation on the grid where
$\lambda \in [10^{-3}, 10^{-6}]$ and $\tau \in [4, 6]$ for the vDem dataset. The reported variance is the standard error of the mean. All the other parameters of the network (number of layers, number of epochs, etc...) have been set as reported in the main experimental section. Overall, the variation (within the defined grid) of the hyperparameters produced some minor effects on the quality of the explanations in terms of accuracy and complexity.

% \usepackage{graphicx}
\begin{table}[!ht]
\centering
\caption{Impact of the hyperparameters on the quality of the model and the explanations.}
\label{tab:hyperparam_explanations}
\resizebox{\columnwidth}{!}{%
\begin{tabular}{llll}
\hline
$(\tau, \lambda)$ & Model accuracy   & Explanation accuracy & Explanation complexity \\ \hline
$(4, 10^{-6})$      & $94.15 \pm 0.53$ & $89.68 \pm 0.36$     & $1.70 \pm 0.20$        \\
$(4, 10^{-5})$      & $94.56 \pm 0.28$ & $88.20 \pm 1.05$     & $2.50 \pm 0.76$        \\
$(4, 10^{-4})$     & $94.53 \pm 0.42$ & $87.73 \pm 1.25$     & $3.10 \pm 0.83$        \\
$(4, 10^{-3})$      & $94.62 \pm 0.45$ & $89.37 \pm 0.73$     & $2.00 \pm 0.16$        \\
$(5, 10^{-6})$      & $94.53 \pm 0.65$ & $89.91 \pm 0.59$     & $2.50 \pm 0.50$        \\
$(5, 10^{-5})$      & $94.86 \pm 0.23$ & $88.04 \pm 1.29$     & $1.60 \pm 0.19$        \\
$(5, 10^{-4})$     & $94.68 \pm 0.48$ & $88.20 \pm 0.80$     & $3.25 \pm 0.92$        \\
$(5, 10^{-3})$      & $94.53 \pm 0.58$ & $89.92 \pm 0.39$     & $2.00 \pm 0.22$        \\
$(6, 10^{-6})$      & $94.83 \pm 0.49$ & $85.53 \pm 1.98$     & $3.30 \pm 0.64$        \\
$(6, 10^{-5})$      & $94.56 \pm 0.51$ & $87.90 \pm 0.98$     & $2.70 \pm 0.58$        \\
$(6, 10^{-4})$     & $94.48 \pm 0.73$ & $87.22 \pm 1.80$     & $1.80 \pm 0.25$        \\
$(6, 10^{-3})$      & $94.05 \pm 0.58$ & $90.15 \pm 0.68$     & $2.12 \pm 0.43$        \\ \hline
\end{tabular}%
}
\end{table}

% \clearpage

\subsection{Logic formulas}
\label{appendix:formulas}
Table \ref{tab:formula_comparison} reports a selection of the rule extracted by each method in all the experiments presented in the main paper. For all methods we report only the explanations of the first class for the first split of the Cross-validation.
At last, for the Entropy-based method only, Tables \ref{tab:mimic_rules}, \ref{tab:vdem_rules}, \ref{tab:mnist_rules}, \ref{tab:cub_rules} resume the explanations of all classes in all experiments. For simplicity, in the following tables we dropped the universal quantifier for all formulas.

\onecolumn
{\renewcommand{\arraystretch}{2.8}

\begin{table}[t]
\caption{Comparison of the formulas obtained in the first run of each experiment for all methods. We dropped the arguments in the logic predicates as well as the universal quantifiers for simplicity. Only the formula explaining the first class has been reported. Ellipses are used to truncate overly long formulas.}
\label{tab:formula_comparison}
\resizebox{\textwidth}{!}{

\begin{tabular}{lll}
\toprule
    \centering
    Dataset &      Method & Formulas \\
\midrule

  MIMIC-II &  Entropy &  
  \makecell[l]{recover  $\leftrightarrow$  $\neg$liver\_flg  $\wedge$  $\neg$stroke\_flg  $\wedge$  $\neg$mal\_flg 
  }\\
   &       DTree &  
  \makecell[l]{recover  $\leftrightarrow$ (age\_high \textless 0.5  $\wedge$  mal\_flg \textless 0.5  $\wedge$  stroke\_flg \textless 0.5  $\wedge$ \\
  age\_normal \textless 0.5  $\wedge$ iv\_day\_1\_normal \textless 0.5) $\vee$ (age\_high \textless 0.5  $\wedge$  mal\_flg \textless 0.5  $\wedge$ \\
  stroke\_flg \textless 0.5  $\wedge$  age\_normal \textless 0.5  $\wedge$  iv\_day\_1\_normal \textgreater 0.5) $\vee$ (age\_high \textless 0.5  $\wedge$ ...\\
%   mal\_flg \textless 0.5  $\wedge$  stroke\_flg \textless 0.5  $\wedge$  age\_normal \textgreater 0.5  $\wedge$  icu\_los\_day\_high \textless 0.5) $\vee$ ... 
%   (age\_high \textless 0.5  $\wedge$  mal\_flg \textless 0.5  $\wedge$  stroke\_flg \textgreater 0.5  $\wedge$  afib\_flg \textless 0.5  $\wedge$  aline\_flg \textless 0.5) $\vee$ (age\_high \textless 0.5  $\wedge$  mal\_flg \textless 0.5  $\wedge$  stroke\_flg \textgreater 0.5  $\wedge$  afib\_flg \textless 0.5  $\wedge$  aline\_flg \textgreater 0.5) $\vee$ (age\_high \textless 0.5  $\wedge$  mal\_flg \textgreater 0.5  $\wedge$  weight\_first\_normal \textless 0.5  $\wedge$  service\_num\_low \textless 0.5  $\wedge$  liver\_flg \textless 0.5) $\vee$ (age\_high \textless 0.5  $\wedge$  mal\_flg \textgreater 0.5  $\wedge$  weight\_first\_normal \textless 0.5  $\wedge$  service\_num... 
  }\\
\bottomrule
\end{tabular}
}
\end{table}
}

\clearpage

{\renewcommand{\arraystretch}{1.5}
\begin{table}[h]
\centering
\caption{Formulas extracted from the MIMIC-II dataset.}
\label{tab:mimic_rules}
\begin{tabular}{p{\textwidth}}
\hline
Formulas \\ \hline
recover $\leftrightarrow$ $\neg$liver\_flg $\wedge$ $\neg$stroke\_flg $\wedge$ $\neg$mal\_flg \\
non\_recover $\leftrightarrow$ mal\_flg $\vee$ (age\_HIGH $\wedge$ $\neg$iv\_day\_1\_NORMAL) \\ \hline
\end{tabular}
\end{table}
}

% \usepackage{longtable}
% Note: It may be necessary to compile the document several times to get a multi-page table to line up properly
{\renewcommand{\arraystretch}{1.5}
\begin{table}[ht]
\centering
\caption{Formulas extracted from the V-Dem dataset.}
\label{tab:vdem_rules}
\begin{tabular}{p{\textwidth}}
\hline
Formulas \\ \hline
non\_electoral\_democracy $\leftrightarrow$ $\neg$v2xel\_frefair $\vee$ $\neg$v2x\_elecoff $\vee$ $\neg$v2x\_cspart $\vee$ $\neg$v2xeg\_eqaccess $\vee$ $\neg$v2xeg\_eqdr \\
electoral\_democracy $\leftrightarrow$ v2xel\_frefair $\wedge$ v2x\_elecoff $\wedge$ v2x\_cspart \\ \hline
\end{tabular}
\end{table}
}

% \usepackage{longtable}
% Note: It may be necessary to compile the document several times to get a multi-page table to line up properly
{\renewcommand{\arraystretch}{1.5}
\begin{table}[ht]
\centering
\caption{Formulas extracted from the MNIST dataset.}
\label{tab:mnist_rules}

\begin{tabular}{p{\textwidth}}
\hline
Formulas \\ \hline
even $\leftrightarrow$ (zero $\wedge$ $\neg$one $\wedge$ $\neg$two $\wedge$ $\neg$three $\wedge$ $\neg$four $\wedge$ $\neg$five $\wedge$ $\neg$six $\wedge$ $\neg$seven $\wedge$ $\neg$eight $\wedge$ $\neg$nine) $\vee$ (two $\wedge$ $\neg$zero $\wedge$ $\neg$one $\wedge$ $\neg$three $\wedge$ $\neg$four $\wedge$ $\neg$five $\wedge$ $\neg$six $\wedge$ $\neg$seven $\wedge$ $\neg$eight $\wedge$ $\neg$nine) $\vee$ (four $\wedge$ $\neg$zero $\wedge$ $\neg$one $\wedge$ $\neg$two $\wedge$ $\neg$three $\wedge$ $\neg$five $\wedge$ $\neg$six $\wedge$ $\neg$seven $\wedge$ $\neg$eight $\wedge$ $\neg$nine) $\vee$ (six $\wedge$ $\neg$zero $\wedge$ $\neg$one $\wedge$ $\neg$two $\wedge$ $\neg$three $\wedge$ $\neg$four $\wedge$ $\neg$five $\wedge$ $\neg$seven $\wedge$ $\neg$eight $\wedge$ $\neg$nine) $\vee$ (eight $\wedge$ $\neg$zero $\wedge$ $\neg$one $\wedge$ $\neg$two $\wedge$ $\neg$three $\wedge$ $\neg$four $\wedge$ $\neg$five $\wedge$ $\neg$six $\wedge$ $\neg$seven $\wedge$ $\neg$nine) \\
\hline
\end{tabular}
\end{table}
}

\clearpage

\onecolumn

% \usepackage{longtable}
% Note: It may be necessary to compile the document several times to get a multi-page table to line up properly
{\renewcommand{\arraystretch}{1.5}
\begin{longtable}[c]{p{\textwidth}}
\caption{Formulas extracted from the CUB dataset.}
\label{tab:cub_rules}\\
\hline
Formulas \\ \hline
\endfirsthead
\multicolumn{1}{c}%
{{\bfseries Table \thetable\ continued from previous page}} \\
\hline
Formulas \\ \hline
\endhead
\hline
\endfoot
\endlastfoot
Black\_footed\_Albatross $\leftrightarrow$ has\_bill\_length\_about\_the\_same\_as\_head $\wedge$ has\_wing\_pattern\_solid $\wedge$ $\neg$has\_upper\_tail\_color\_grey $\wedge$ $\neg$has\_belly\_color\_white $\wedge$ $\neg$has\_wing\_shape\_roundedwings $\wedge$ $\neg$has\_bill\_color\_black \\
Laysan\_Albatross $\leftrightarrow$ has\_crown\_color\_white $\wedge$ has\_wing\_pattern\_solid $\wedge$ $\neg$has\_under\_tail\_color\_white \\
Sooty\_Albatross $\leftrightarrow$ has\_upper\_tail\_color\_grey $\wedge$ has\_size\_medium\_9\_\_16\_in $\wedge$ has\_bill\_color\_black $\wedge$ $\neg$has\_belly\_color\_white \\
Groove\_billed\_Ani $\leftrightarrow$ has\_breast\_color\_black $\wedge$ has\_leg\_color\_black $\wedge$ $\neg$has\_bill\_shape\_allpurpose $\wedge$ $\neg$has\_bill\_length\_about\_the\_same\_as\_head $\wedge$ $\neg$has\_wing\_shape\_roundedwings \\
Crested\_Auklet $\leftrightarrow$ has\_nape\_color\_black $\wedge$ $\neg$has\_eye\_color\_black $\wedge$ $\neg$has\_belly\_color\_white \\
Least\_Auklet $\leftrightarrow$ has\_breast\_color\_black $\wedge$ has\_breast\_color\_white $\wedge$ $\neg$has\_nape\_color\_white $\wedge$ $\neg$has\_size\_small\_5\_\_9\_in \\
Parakeet\_Auklet $\leftrightarrow$ has\_size\_medium\_9\_\_16\_in $\wedge$ has\_primary\_color\_white $\wedge$ has\_leg\_color\_grey \\
Rhinoceros\_Auklet $\leftrightarrow$ has\_size\_medium\_9\_\_16\_in $\wedge$ has\_leg\_color\_buff \\
\hline
\end{longtable}
}

\end{document}